\documentclass{winnower}
\pdfoutput=1

\hypersetup{
    colorlinks = true,
	citecolor=blue,
	linkcolor=blue
}

\usepackage{enumerate}
\usepackage{algorithm,algorithmic}
\usepackage{amsmath,amssymb}
\usepackage{wrapfig}
\usepackage{multirow}
\usepackage{booktabs}

\usepackage{comment}
\usepackage{natbib}

\usepackage{subfigure}

\newcommand{\best}[1]{\textbf{#1}}
\newcommand{\second}[1]{\underline{#1}}

\usepackage{algorithmic,algorithm}
 \usepackage{setspace}

\begin{document}

\title{Stochastic Neighbor Embedding of Multimodal Relational Data \\ for Image-Text Simultaneous Visualization}

\author[1]{Morihiro Mizutani\thanks{mizutani@sys.i.kyoto-u.ac.jp}}
\author[2]{Akifumi Okuno\thanks{oknakfm@gmail.com}}
\author[3]{Geewook Kim\thanks{gwkim.rsrch@gmail.com}\footnote{Work done while at Kyoto University and RIKEN Center for AIP.}}
\author[1,2]{Hidetoshi Shimodaira\thanks{shimo@i.kyoto-u.ac.jp}}
\affil[1]{Graduate School of Informatics, Kyoto University}
\affil[2]{RIKEN Center for Advanced Intelligence Project~(AIP)}
\affil[3]{NAVER Corporation}

\date{}

\maketitle

\begin{abstract}

Multimodal relational data analysis has become of increasing importance in recent years, for exploring across different domains of data, such as images and their text tags obtained from social networking services~(e.g., Flickr). 
A variety of data analysis methods have been developed for visualization; to give an example, $t$-Stochastic Neighbor Embedding ($t$-SNE) computes low-dimensional feature vectors so that their similarities keep those of the observed data vectors. 
However, $t$-SNE is designed only for a single domain of data but not for multimodal data; this paper aims at visualizing multimodal relational data consisting of data vectors in multiple domains with relations across these vectors. 
By extending $t$-SNE, we herein propose \emph{Multimodal Relational Stochastic Neighbor Embedding (MR-SNE)}, that (1) first computes augmented relations, where we observe the relations across domains and compute those within each of domains via the observed data vectors, and (2) jointly embeds the augmented relations to a low-dimensional space. Through visualization of Flickr and Animal with Attributes~2 datasets, 
proposed MR-SNE is compared with other graph embedding-based approaches; MR-SNE demonstrates the promising performance.
\end{abstract}

\section{Introduction}
\label{sec:intro}

Many different types of data called \emph{multimodal data}~(i.e., text, image, and audio) has become readily available over recent years, by virtue of the rapid development of information technology. 
The data type is especially called \emph{domain} or \emph{view}, and analyzing the relations across different domains has attracted extensive attention~\citep{davidson2010youtube,sharma2012generalized,hong2014image,mathew2016book,chu2017hybrid}.

As can be easily assumed from the surrounding environment of image processing, 
datasets employed recently 
comprise high-dimensional data vectors and their complicated relations. 
For instance, images and their relevant texts have complicated graph-structured relations, as several texts are simultaneously associated with each of the images. 
Such a relation across domains is characterized by an \emph{across-domains graph}; 
data vectors in each domain and the across-domains graph are together called \emph{multimodal relational data}.

However, these data vectors are perplexing to comprehend by humans due to their high-dimensionality. 
Thus \emph{data visualization} has an indispensable role in delving into the observed relational data, as it converts the perplexing high-dimensional data vectors into the corresponding low-dimensional vectors that are easily comprehensible to humans. Their underlying relations within and across domains are then expected to be identified with external human knowledge.

One of the most popular approaches for visualization in recent years is $t$-stochastic neighbor embedding~\citep[$t$-SNE;][]{t-SNE}, that first computes similarities between all pairs of the data vectors, where they are especially called \emph{stochastic neighbor graph}, and it subsequently computes the low-dimensional vectors so that their graph is preserved. 
$t$-SNE is classified into manifold learning~\citep{LE,cayton2005algorithms,talwalkar2008large}, which is well developed so far.  
However, manifold learning requires defining similarities between data vectors; \textbf{$t$-SNE cannot be applied to multimodal data}, that contains data vectors whose different dimensionalities depend on the domain.

\begin{figure}[htbp]
\centering
\subfigure[CDMCA~(0.657)]{
		\includegraphics[width=3.3cm]{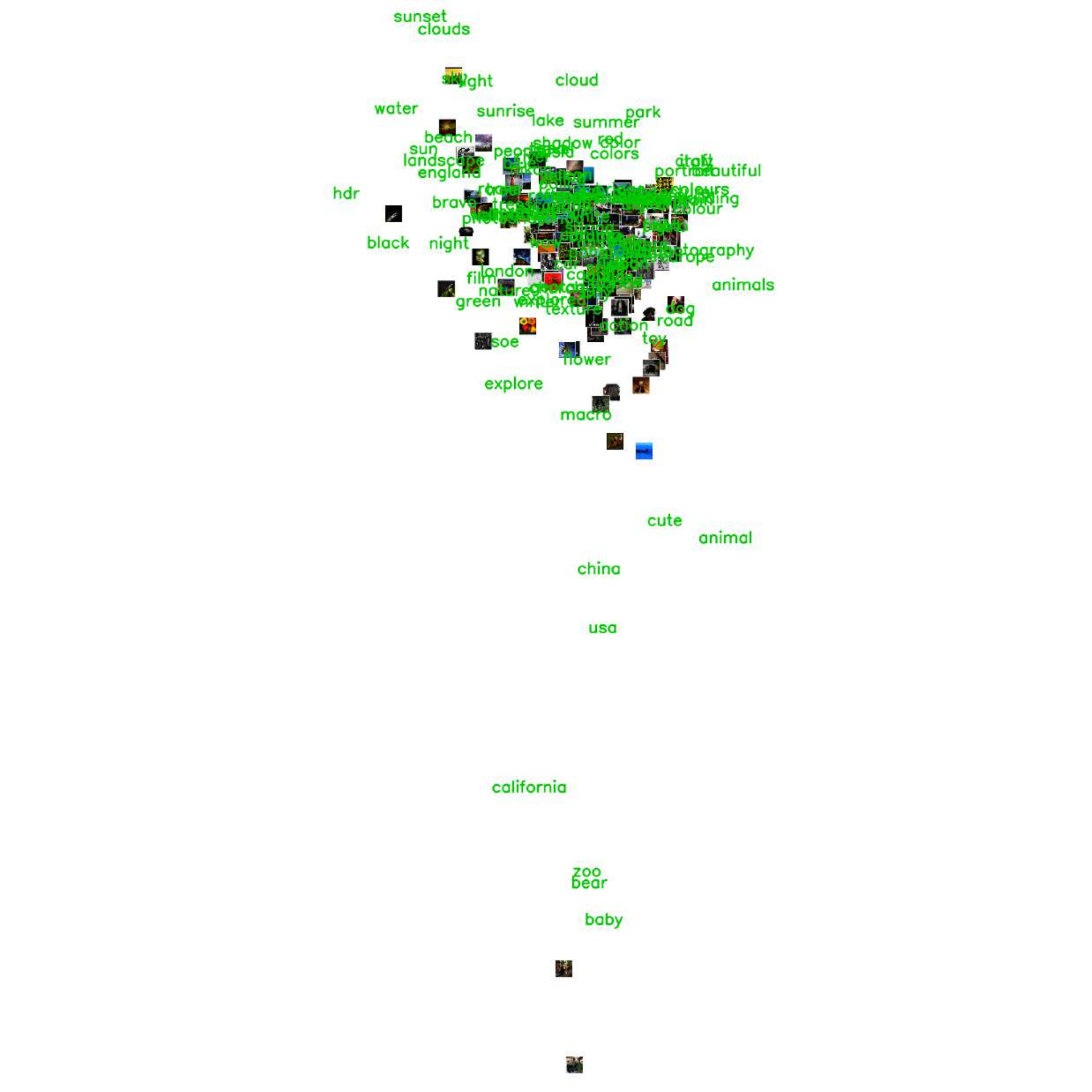}
	\label{fig:cdmca}
}
\hspace{1em}
\subfigure[PMvGE~(0.811) \protect\newline Inner-Product]{
		\includegraphics[width=3.3cm]{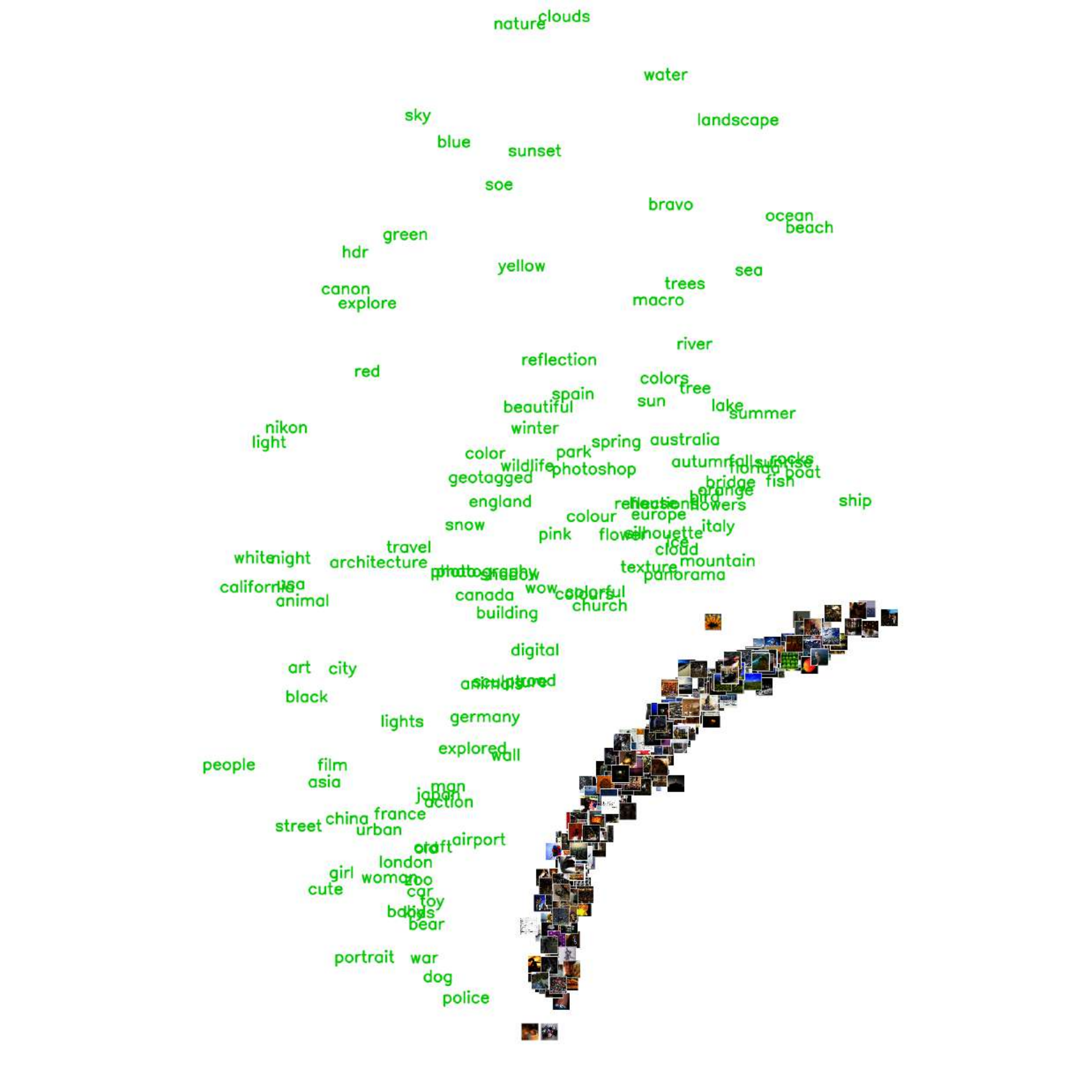}
	\label{fig:pmvge}
}
\hspace{1em}
\subfigure[PMvGE~(0.884) \protect\newline Squared-Distance]{
		\includegraphics[width=3.3cm]{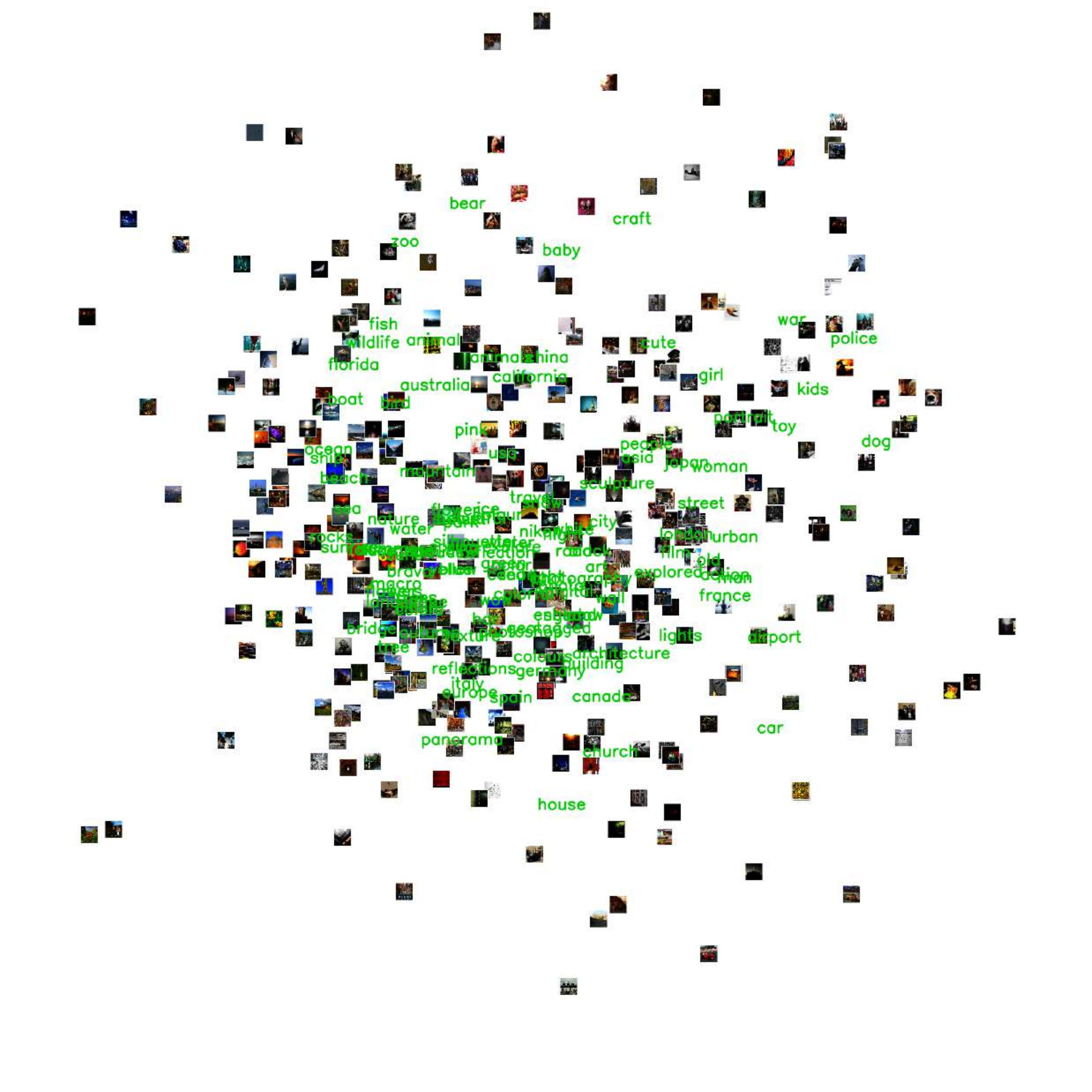}
	\label{fig:pmvge_euclidean}
}
\hspace{1em}
\subfigure[\textbf{MR-SNE}~(0.828) \protect\newline Proposal]{
\includegraphics[width=3.3cm]{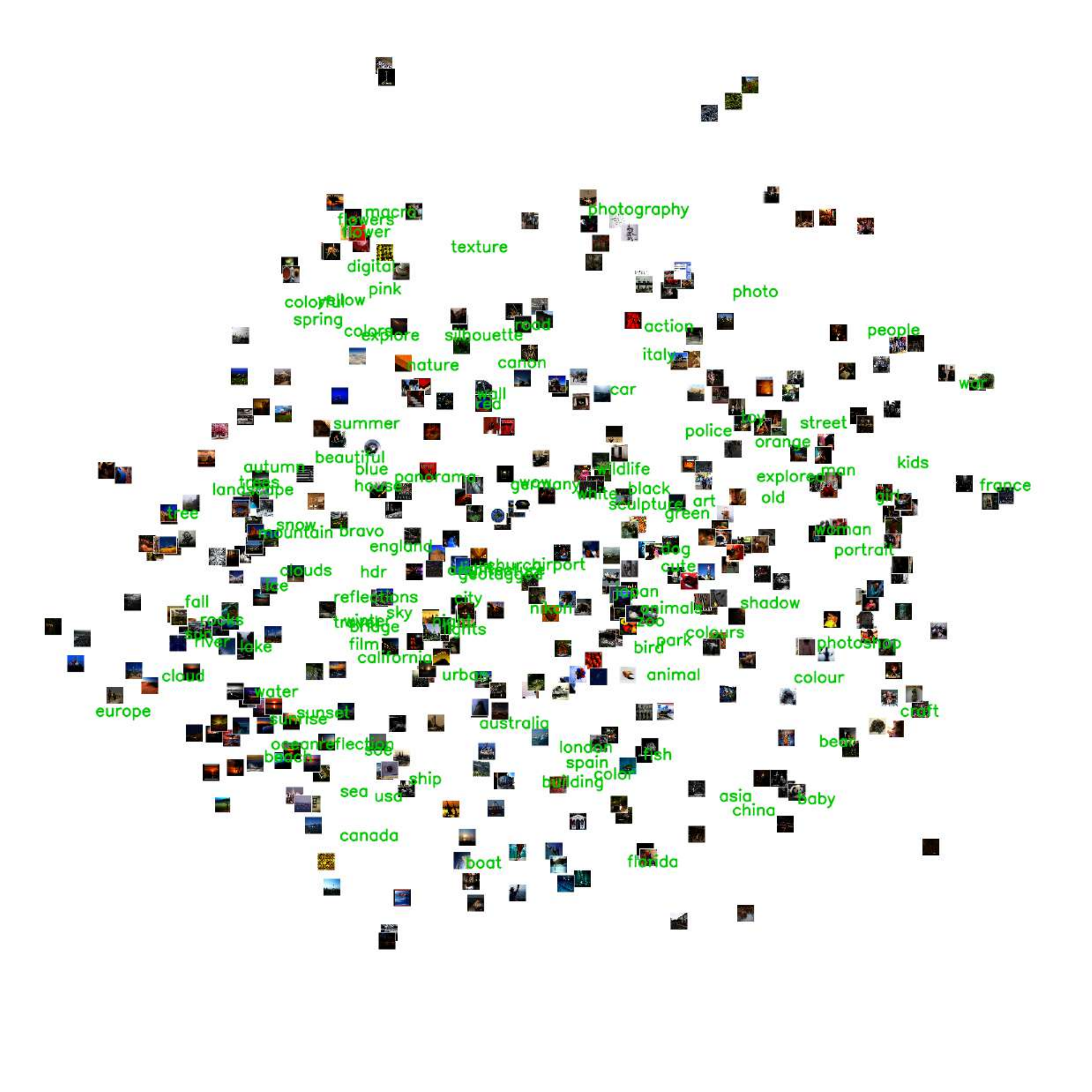}
	\label{fig:proposal_intro}
}
\caption{An observed graph across text and image domains in the Flickr dataset is embedded into a $2$-dim. subspace. ROC-AUC scores for link prediction across domains are provided in parentheses~(higher is better). 
Note that the existing $t$-SNE is not listed here as it cannot be applied to the multimodal relational data. 
See Section~\ref{sec:experiments} for further details.}
\label{fig:comparison_intro}
\end{figure}

Although $t$-SNE cannot be directly applied, there have been a variety of alternatives for visualizing multimodal relational data. 
One way is to employ cross-domain matching correlation analysis~\citep[CDMCA;][Figure~\ref{fig:cdmca}]{CDMCA}, which incorporates graph embedding into linear canonical correlation analysis~\citep[CCA;][]{CCA} so that the low-dimensional vectors represent observed graph-structured relations. 
Probabilistic multi-view graph embedding~\citep[PMvGE;][]{PMvGE} employs highly expressive neural networks for transforming data vectors to low-dimensional feature vectors. 
PMvGE predicts an underlying relation via similarity between the pair of obtained feature vectors, which may provide a good visualization.
In our link prediction experiment across domains, PMvGE using inner-product similarity and squared Euclidean distance demonstrate a better ROC-AUC score than CDMCA and random prediction.

However, PMvGE with only the observed across-domains graph \textbf{does not fully leverage the underlying but unobserved relations within each of domains}, that are also crucial for visualization;
its visualization can be unsatisfactory in several practical cases. 
For instance, Figure~\ref{fig:pmvge} shows that the text and image feature vectors obtained via PMvGE with inner-product similarity are completely separated, even though it shows a good ROC-AUC score for link prediction across domains. 
\citet{mimno-thompson-2017-strange} reported the same phenomenon in skip-gram~\citep{word2vec}~(a.k.a. word2vec), that similarly considers only a graph across word and context domains. 
Figure~\ref{fig:pmvge_euclidean} shows that text vectors are unnaturally concentrated around the center when squared Euclidean distance is employed.

\begin{wrapfigure}[20]{l}[0pt]{7cm}
\centering
\vspace{0.5em}
\includegraphics[width=5.2cm]{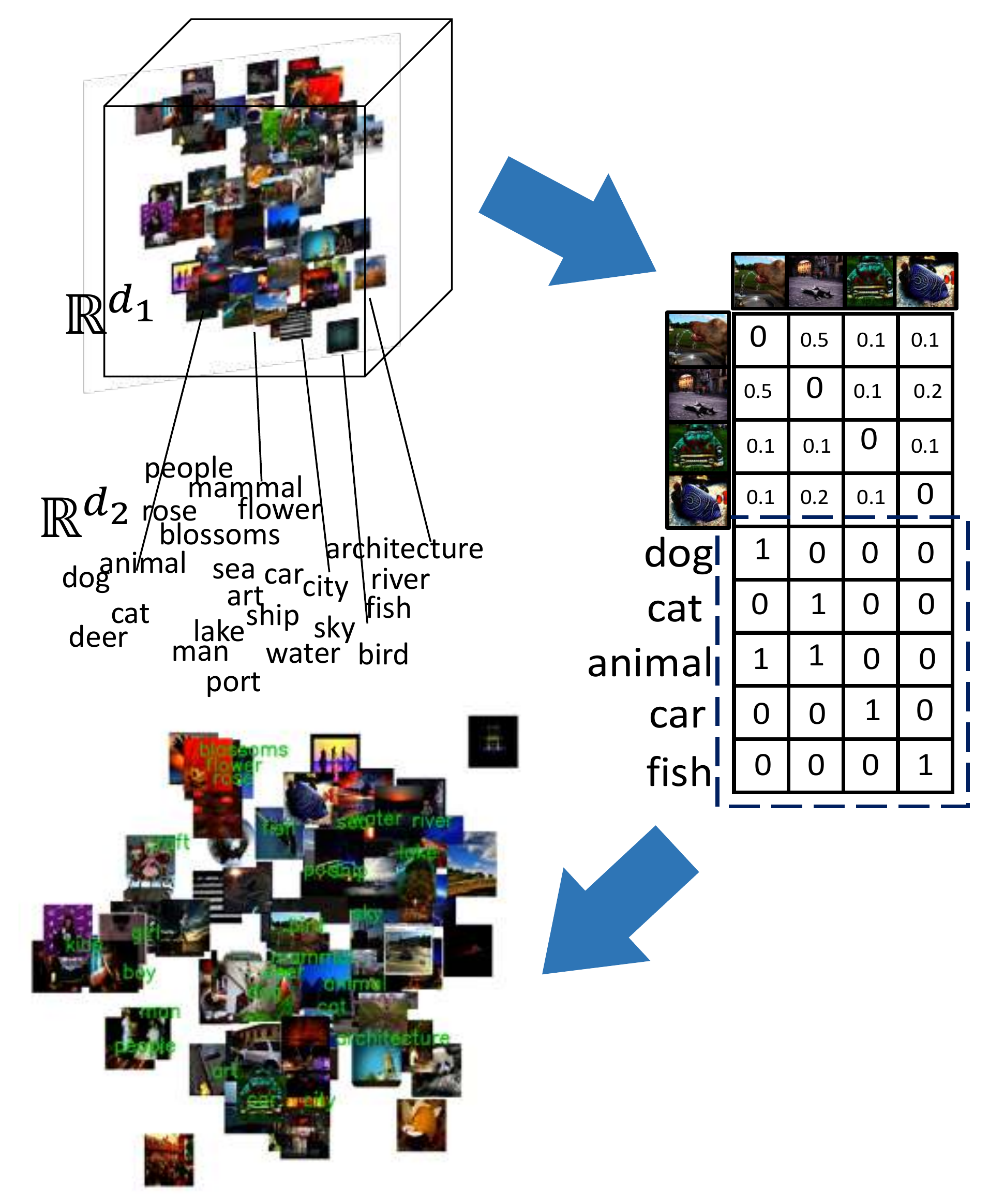}
\caption{Proposed MR-SNE: Observed across-domains graph and stochastic neighbor graphs computed within each of domains are jointly embedded into the common subspace, by applying $t$-SNE framework.}
\label{fig:proposal}
\end{wrapfigure}

Therefore, this paper extends $t$-SNE to the multimodal setting, so that not only the observed across-domains graph but also stochastic neighbor graphs computed within each domain are considered, as shown in Figure~\ref{fig:proposal}.

\newpage

\begin{figure*}
\centering
\includegraphics[width=0.80\textwidth]{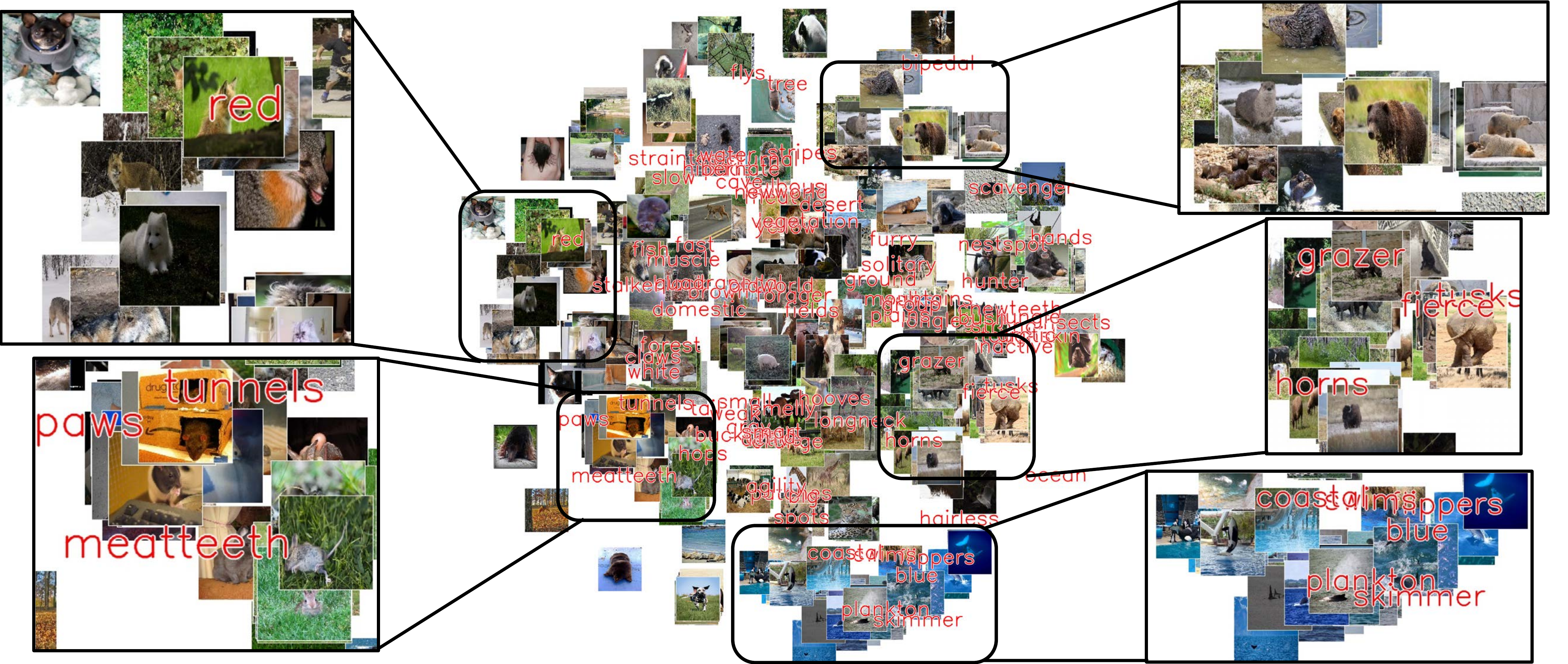}
\caption{Simultaneous visualization of $n_1=2500$ images and their $n_2=85$ text tags in Animal with Attributes 2 dataset~\citep{AWA2}.
Randomly selected 500 images are shown here.
The data vectors are  $d_1=2048$ dim.\ CNN image vectors $\{x^{(1)}_i\}_{i=1}^{n_1}$ and $d_2=300$ dim.\ GloVe word vectors $\{x^{(2)}_j\}_{j=1}^{n_2}$. 
These data vectors are embedded in $K=2$-dim.\ common subspace, by applying the proposed MR-SNE. See Fig.~\ref{fig:visualization_Flickr} in Supplementary Material~\ref{supp:visualization_Flickr} for Flickr dataset.
}
\label{fig:visualization_AWA2}
\end{figure*}

\textbf{Our contribution:} 
We propose \emph{Multimodal Relational Stochastic Neighbor Embedding (MR-SNE)}, that (i) first computes stochastic neighbor graphs within each domain so that they predict the underlying but unobserved relations, 
(ii) computes augmented relations for the multimodal relational data, by linking up the observed across-domains graph and the stochastic neighbor graphs computed within each of domains, and 
(iii) jointly embeds the augmented relations by applying the existing $t$-SNE framework.
The proposed MR-SNE extends $t$-SNE to the multimodal setting, and it provides a good visualization, as shown in Figure~\ref{fig:visualization_AWA2}. 
Extensive numerical experiments are conducted through visualization of Flicker and Animal with Attributes~2 (AwA2) dataset; other possible approaches for visualizing multimodal relational data are also empirically compared with the proposed MR-SNE. 

\vspace{0.5em}
\textbf{Organization of this paper:} problem setting is summarized in Section~\ref{sec:problem_setting}, 
proposed MR-SNE is described in Section~\ref{sec:proposed_method}, numerical experiments are conducted in Section~\ref{sec:experiments}, and Section~\ref{sec:conclusion} concludes this paper.

\vspace{2em}

\subsection{Related Works}
\label{subsec:related_works}
Laplacian eigenmaps~\citep{LE}, multidimensional scaling~\citep[MDS;][]{MDS} and Isomap~\citep{Isomap} are manifold learning methods; they cannot be applied to multimodal data, similarly to $t$-SNE~\citep{t-SNE} and principal component analysis~\citep[PCA;][]{PCA}.
Multi-view SNE~\citep[m-SNE;][]{xie2011msne} considers vector representations of objects in $2$ domains, indicating their $2$ different perspectives; the objects therein are essentially $1$-view, and m-SNE consequently provides one vector representation for each object by aggregating these $2$-domains. Thus the problem setting is different from this paper.
One-to-one relations across $2$ domains and stochastic neighbor graphs computed within each of $2$ domains are simultaneously considered in \citet{quadrianto2011learning}, \citet{sun2007locality}, Zhao and Evans-Dugelay~\cite{zhao2013unsupervised} and \citet{wang2013new}, but they cannot be applied to multimodal relational data.

\section{Problem Setting}
\label{sec:problem_setting}

In this section, we describe the problem setting when $D=2$ domains are considered. It can be straightforwardly generalized to arbitrary $D \ge 1$. 

Let $d_1,d_2,n_1,n_2 \in \mathbb{N}$ and let $\{x_i^{(1)}\}_{i=1}^{n_1} \subset \mathbb{R}^{d_1},\{x_j^{(2)}\}_{j=1}^{n_2} \subset \mathbb{R}^{d_2}$ be data vectors for the two domains. 
$W=(w_{ij}) \in \mathbb{R}_{\ge 0}^{n_1 \times n_2}$ is the observed across-domains graph; $w_{ij} \ge 0$ represents the strength of relation between $x^{(1)}_i \in \mathbb{R}^{d_1}$ and $x^{(2)}_j \in \mathbb{R}^{d_2}$, where $w_{ij}=0$ represents no relation. 
Then, our goal is to provide a better visualization of the multimodal relational data $(\{x^{(1)}_i\}_{i=1}^{n_1},\{x^{(2)}_j\}_{j=1}^{n_2},\{w_{ij}\}_{i,j=1}^{n_1,n_2})$, by transforming $d_1,d_2$-dimensional data vectors in each domain into the user-specified $K( \le \min\{d_1,d_2\})$-dimensional feature vectors $\{y_i^{(1)}\}_{i=1}^{n_1},\{y_j^{(2)}\}_{j=1}^{n_2} \subset \mathbb{R}^K$. 
Particularly, $K=2$ is considered for visualization.

\textbf{A typical example} is the simultaneous visualization of images and their text tags, which is shown in Figure~\ref{fig:visualization_AWA2}. 
In this example, (i) data vectors for images $\{x_i^{(1)}\}_{i=1}^{n_1} \subset \mathbb{R}^{d_1}$ and their text tags $\{x^{(2)}_j\}_{j=1}^{n_2} \subset \mathbb{R}^{d_2}$, and (ii) relations $\{w_{ij}\}_{i,j=1}^{n_1,n_2}$ across image and text domains, are observed; $w_{ij}=1$ indicates that the image $i$ is tagged with the text $j$ and $w_{ij}=0$ otherwise.

\textbf{Current issue} is that the conventional $t$-SNE cannot be applied to multimodal relational data, whose dimensions $d_1,d_2$ of $2$ domains are different, as it requires measuring Euclidean distance between data vectors. 
For instance, considering the case that $d_1=10,d_2=20$, Euclidean distance between $x^{(1)}_i \in \mathbb{R}^{10},x^{(2)}_j \in \mathbb{R}^{20}$ is not well-defined.

\section{Proposed Method}
\label{sec:proposed_method}

In this section, we propose an extension of $t$-SNE, that can be applied to multimodal relational data. 
Stochastic neighbor graph is defined in Section~\ref{subsec:stochastic_neighbor}, and 
the proposed method is described in Section~\ref{subsec:proposal}. 
Although the case of $D=2$ domains is considered, it can be straightforwardly extended to general $D \ge 1$; 
$D=1$ corresponds to the conventional $t$-SNE.

\subsection{Stochastic neighbor Graph Within Each Domain}
\label{subsec:stochastic_neighbor}

\begin{wrapfigure}[12]{r}[0pt]{4cm}
\centering
\vspace{-1.5em}
\includegraphics[width=3.6cm]{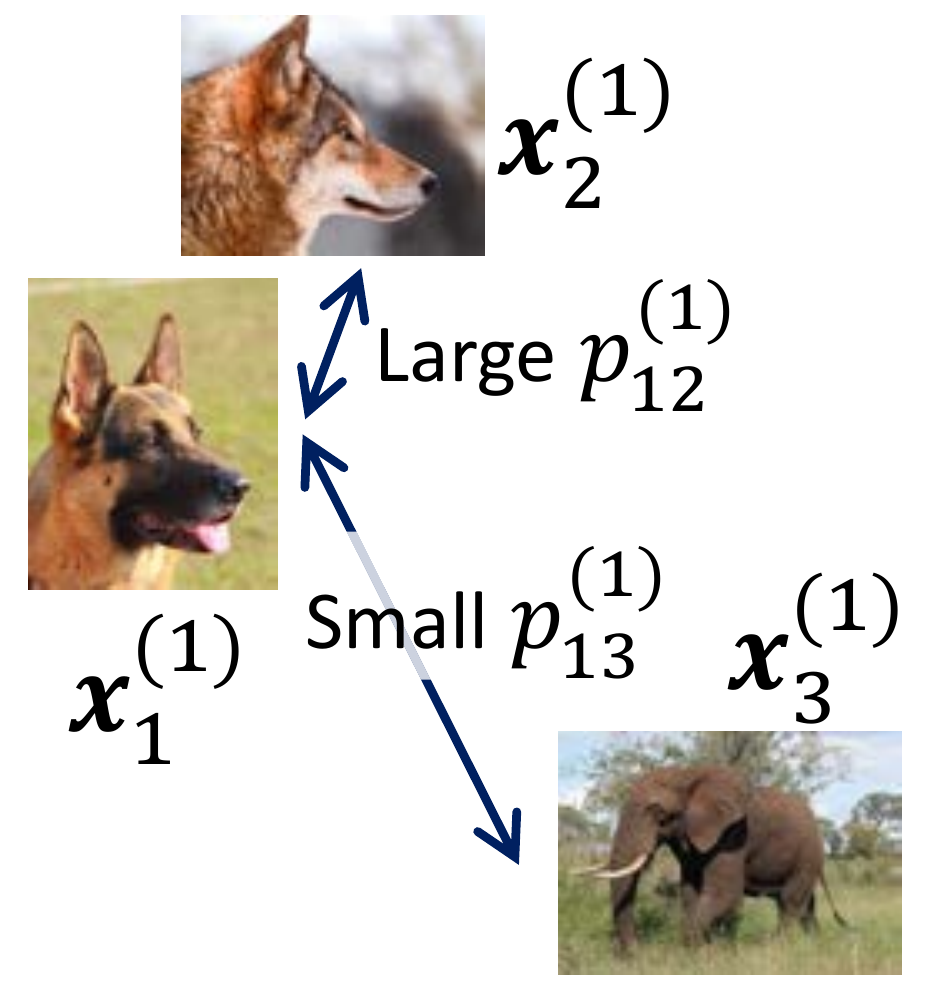}
\vspace{-0.5em}
\caption{$p^{(1)}_{ij}$ is computed depending on the distance between vectors.}
\label{fig:stochastic_neightbor_graph}
\end{wrapfigure}

In this section, we explain the stochastic neighbor graph within each domain, that is defined in \citet{t-SNE} for the case $D=1$. 
Firstly, let us think of the graph for domain $1$. 

Conditional probability for each pair $(x^{(1)}_i,x^{(1)}_j)$ is defined as
\begin{align}
p_{j|i}^{(1)}
&= 
\scalebox{0.85}{$\displaystyle
\frac{{\rm exp}(-\|x^{(1)}_i - x^{(1)}_j \|_2^2/2s^2_i)}{\sum_{k \ne i} \ \exp(-\|x^{(1)}_i - x^{(1)}_k \|_2^2/2s^2_i)}$},
\label{eq:tSNE_pij}
\end{align}
where $s_i>0 \: (i=1,2,\ldots,n_1)$ are  user-specified parameters. 
In \citet{t-SNE}, the parameters $\{s_i\}_{i=1}^{n_1}$ are determined by (i) specifying only one parameter $\lambda>0$ called ``perplexity", and (ii) solving equations $\lambda=2^{H_i}$ with $H_i:=-\sum_{j}p^{(1)}_{j \mid i} \log p^{(1)}_{j \mid i}$, for $i=1,2,\ldots,n$.
Using (\ref{eq:tSNE_pij}), pairwise similarity is defined as $p^{(1)}_{ij}:=(p^{(1)}_{i \mid j} + p^{(1)}_{j \mid i}) / 2n_1$ for $i \ne j$ and $p^{(1)}_{ii}=0$; 
\textbf{stochastic Neighbor graph} is defined as $P^{(1)}=(p^{(1)}_{ij})$. 
$t$-SNE regards each entry $p^{(1)}_{ij}$ as a joint probability of the pair $(x^{(1)}_i,x^{(1)}_j)$, as $\{p^{(1)}_{ij}\}$ are non-negative and their total sum is $1$.
The graph $P^{(2)}$ for domain $2$ is similarly defined.

\subsection{Proposed Method}
\label{subsec:proposal}

In this section, we propose Multimodal Relational Stochastic Neighbor Embedding (MR-SNE) that visualizes the multimodal relational data, as shown in Figure~\ref{fig:proposal}; the procedure is formally defined in the following (i)-(iii), and is also shown in Algorithm~\ref{alg:proposal} in Supplement~\ref{supp:algorithm}.

\begin{enumerate}[{(i)}]
\item[(i-1)] Stochastic neighbor graphs $P^{(1)}=(p_{ij}^{(1)}) \in \mathbb{R}_{\ge0}^{n_1\times n_1}, P^{(2)}=(p_{ij}^{(2)}) \in \mathbb{R}_{\ge0}^{n_2\times n_2}$ for each domain $1,2$ are computed as explained in Section~\ref{subsec:stochastic_neighbor}.

\item[(i-2)] 
Observed across-domains graph $W=(w_{ij}) \in \mathbb{R}^{n_1 \times n_2}_{\ge 0}$ is normalized to $r_{ij}
=
\frac{w_{ij}}{\sum_{i=1}^{n_1}\sum_{j=1}^{n_2}w_{ij}}$. 
$r_{ij}$ is regarded as the joint probability of the pair of $2$ domain vectors $(x^{(1)}_i,x^{(2)}_j)$. 
We denote the normalized across-domains graph by $R=(r_{ij}) \in \mathbb{R}^{n_1 \times n_2}_{\ge 0}$.

\item[(i-3)] 
We construct a matrix $\tilde{P} \in \mathbb{R}_{\ge 0}^{(n_1+n_2)\times(n_1 + n_2)}$ from the above matrices $P^{(1)},P^{(2)},R$ as
\begin{align}
\tilde{P}
:=
\scalebox{0.85}{$\displaystyle
\left( \begin{array}{cc}
    \beta_{1} P^{(1)} & \frac{\beta_{12}}{2}R \\
    \frac{\beta_{12}}{2}R^{\top} & \beta_{2} P^{(2)} \\
\end{array} \right)$}, 
\label{eq:tildeP}
\end{align}
where $\beta_{1},\beta_{12},\beta_{2} \ge 0$ are user-specified parameters satisfying a constraint $\beta_{1}+\beta_{12}+\beta_2=1$. 
Then, we regard the entries in $\tilde{P}$ as a joint probability of the corresponding data vectors, as all the entries are non-negative and their total sum is $1$.

\item[(ii)] 
We consider the corresponding low-dimensional feature vectors $\{y^{(1)}_i\}_{i=1}^{n_1},\{y^{(2)}_i\}_{i=1}^{n_2} \subset \mathbb{R}^K$, and define a joint probability of $(y^{(d)}_{i},y^{(e)}_j)$ for $i \in [n_d], j \in [n_e], d,e \in [2]$ by
\begin{align}
q^{(d,e)}_{ij} 
= 
\scalebox{0.85}{$\displaystyle
\frac{
    (1+\|y^{(d)}_i-y^{(e)}_j\|_2^2)^{-1}
}{
    \sum_{(f,k)} \sum_{(g,l) \ne (f,k)} (1+\|y^{(f)}_k-y^{(g)}_l\|_2^2)^{-1}
} \quad (\text{if }(d,i) \ne (e,j))$},
\end{align}
and $q^{(d,d)}_{ii}=0$. 
We construct a matrix $\tilde{Q}  \in \mathbb{R}_{\ge 0}^{(n_1+n_2)\times(n_1 + n_2)}$ from the matrices $Q^{(1)}=(q^{(1,1)}_{ij}),Q^{(2)}=(q^{(2,2)}_{ij}),Q^{(1,2)}=(q^{(1,2)}_{ij})$ as
\begin{align}
    \tilde{Q}
    :=
    \scalebox{0.85}{$\displaystyle
    \left(\begin{array}{cc}
        Q^{(1)} & Q^{(1,2)} \\
        (Q^{(1,2)})^{\top} & Q^{(2)} \\
    \end{array}\right)$}.
    \label{eq:tildeQ}
\end{align}

\item[(iii)] 
Low-dimensional vectors $\{y^{(1)}_i\}_{i=1}^{n_1},\{y^{(2)}_i\}_{i=1}^
{n_2} \subset \mathbb{R}^K$ are consequently computed by minimizing the Kullback-Leibler divergence between the two distributions $\tilde{P},\tilde{Q}$, that is
\begin{align}
\tilde{C}
&=
\text{KL}(\tilde{P} || \tilde{Q}) 
= 
\sum_{d=1}^{2}
\beta_{d}
\underbrace{
\scalebox{0.85}{$\displaystyle
\sum_{i=1}^{n_d}
\sum_{j=1}^{n_d}
p^{(d)}_{ij} \log \frac{p^{(d)}_{ij}}{q^{(d,d)}_{ij}}
$}
}_{\text{within each domain}} 
+
\beta_{12}
\underbrace{
\scalebox{0.85}{$\displaystyle 
\sum_{i=1}^{n_1} \sum_{j=1}^{n_2} 
r_{ij} \log \frac{r_{ij}}{q^{(1,2)}_{ij}}
$}
}_{\text{across domains}}
+
\text{Const.},
\label{eq:objective_proposal}
\end{align}
where $\beta_1,\beta_{12},\beta_2 \ge 0$ are parameters specified in (\ref{eq:tildeP}). 
In practice, we employ full-batch gradient descent~\citep{GD} with momentum, to optimize the objective function~(\ref{eq:objective_proposal}). 
\end{enumerate}

\section{Numerical Experiments}
\label{sec:experiments}

In this section, we conduct numerical experiments on visualizing $2$-domain data vectors of images and their text tags, in the $(K=)2$-dimensional common subspace.

\begin{itemize}
\item \textbf{Datasets}: 
We employ two datasets consisting of $n_1$ images, $n_2$ text tags, and their graph across domains $W=(w_{ij}) \in \mathbb{R}^{n_1 \times n_2}_{\ge 0}$. $w_{ij}=1$ represents that the text $j$ is tagged with the image $i$, and $w_{ij}=0$ otherwise. Within-domain graphs are not observed.

\begin{itemize}
\item For \textbf{Flickr} dataset~\citep{Flickr}, $n_1=1000$ images are picked up at random, and the most frequently appeared $n_2=613$ text tags are selected. 
$d_1=2048$-dimensional image vectors are computed by CNN using ImageNet~\citep{Inception_v3}. 
We here employ $1$-hot vectors for representing text tags, i.e., $y^{(2)}_j \in \{0,1\}^{n_2}$ whose $j$-th entry is $1$ and $0$ otherwise; distributed representations of words, such as word2vec~\citep{word2vec}, may be employed instead. 

\item For \textbf{Animal with Attributes 2~(AwA2)} dataset~\citep{AWA2}, we employ $n_1=2500$ images consisting of 50 images for each of 50 classes, and their associated $n_2=85$ text tags. $d_1=2048$-dimensional image vectors are computed by ResNet~\citep{ResNet}, and $d_2=300$-dimensional word vectors are computed by GloVe~\citep{glove}. 
\end{itemize}

\item \textbf{Baselines}: 
The following methods are compared to the proposed MR-SNE. 

\begin{itemize}
\item \textbf{CDMCA}~\citep{CDMCA} extends linear CCA~\citep{CCA} so that across-domains graph $W=(w_{ij})$ is considered. 
Kernel CDMCA incorporates Gaussian kernel into CDMCA. They both are also regularized. See Supplement~\ref{supp:cdmca} for details. 
\item \textbf{PMvGE}~\citep{PMvGE} computes feature vectors $y^{(d)}_i:=f^{(d)}_{\theta}(x^{(d)}_i) \in \mathbb{R}^K$ via neural networks $f^{(d)}_{\theta}:\mathbb{R}^{p_d} \to \mathbb{R}^K$ for each domain $d=1,2$, by maximizing the likelihood using a model $w_{ij} \overset{\text{indep.}}{\sim} \text{B}(\sigma(-s(y^{(1)}_i,y^{(2)}_j)))$. 
$\text{B}(\mu)$ represents Bernoulli distribution whose expectation is $\mu \in [0,1]$, and $\sigma(z):=(1+\exp(z))^{-1}$. 
Although arbitrary $s:\mathbb{R}^{2K} \to \mathbb{R}$ can be used, $s(y,y'):=\|y-y'\|_2^2$ is here considered; see Figure~\ref{fig:comparison_intro} for visualization with inner-product $s(y,y')=-\langle y,y' \rangle$.
\end{itemize}

\item \textbf{Preprocessing}: the following options are considered.
\begin{itemize}
    \item For CDMCA / Kernel CDMCA / PMvGE:  (\textbf{ignored}) unobserved within-domain graphs $W^{(1)},W^{(2)}$ are not considered as usual.
    \item For PMvGE:
    (\textbf{2nd-order}) $\hat{W}^{(1)}=W^{\top}W,\hat{W}^{(2)}=WW^{\top}$ called 2nd-order proximity (see, e.g., \citet{tang2015line}) are alternatively employed for domain $1,2$. 
    \item For the proposed MR-SNE, 
    the across-domains graph $W$ is 
    (\textbf{unnorm}) used without normalization, 
    (\textbf{norm}) normalized as $\frac{w_{ij}}{\sqrt{\sum_{k \ne i} w_{ik}}\sqrt{\sum_{l \ne j}w_{lj}}}$, 
    (\textbf{PMI}) normalized as $\frac{w_{ij}}{\sum_{k \ne i} w_{ik}\sum_{l \ne j}w_{lj}}$ similarly to Point-wise Mutual Information~\citep{PMI}.  
\end{itemize}

\item \textbf{Training}: 
The proposed model is trained by fullbatch gradient descent summarized in Algorithm~\ref{alg:proposal} in Supplement~\ref{supp:algorithm}, with the number of iterations $T=500$, learning rate $\eta=100$ and momentum $\alpha=0.5$. 
Also see Supplement~\ref{supp:computational_time} for computational time. 
PMvGE is trained by minibatch SGD with negative sampling, that is used in a variety of existing researches, such as word2vec~\citep{word2vec}, LINE~\citep{tang2015line}.
\end{itemize}

\subsection{Experiment 1: Visualization}
\label{subsec:experiment1}

Flickr and AwA2 datasets are visualized in the following Figures~\ref{fig:Visualization_result_Flickr} and \ref{fig:Visualization_result_AwA}, respectively. 
These plots show 500 images randomly selected from the $n_1$ images.
Also, the 120 most frequently used tags in Flickr and all 85 tags in AwA2 are displayed in green.

\begin{figure}[H]
\centering
\subfigure[CDMCA]{
		\includegraphics[width=3.5cm]{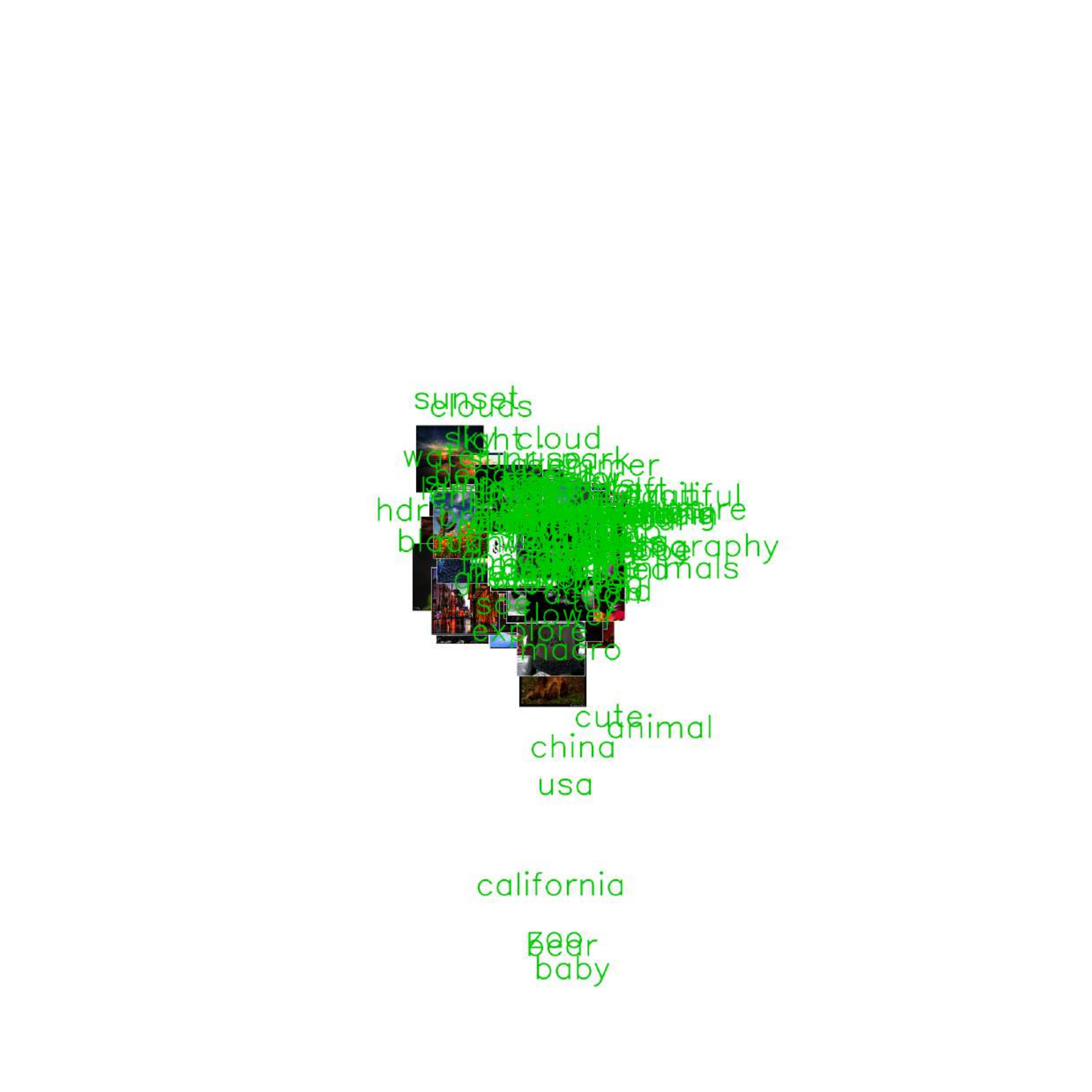}
	\label{fig:Flickr_rCDMCA}
}
\hspace{2em}
\subfigure[Kernel CDMCA]{
		\includegraphics[width=3.5cm]{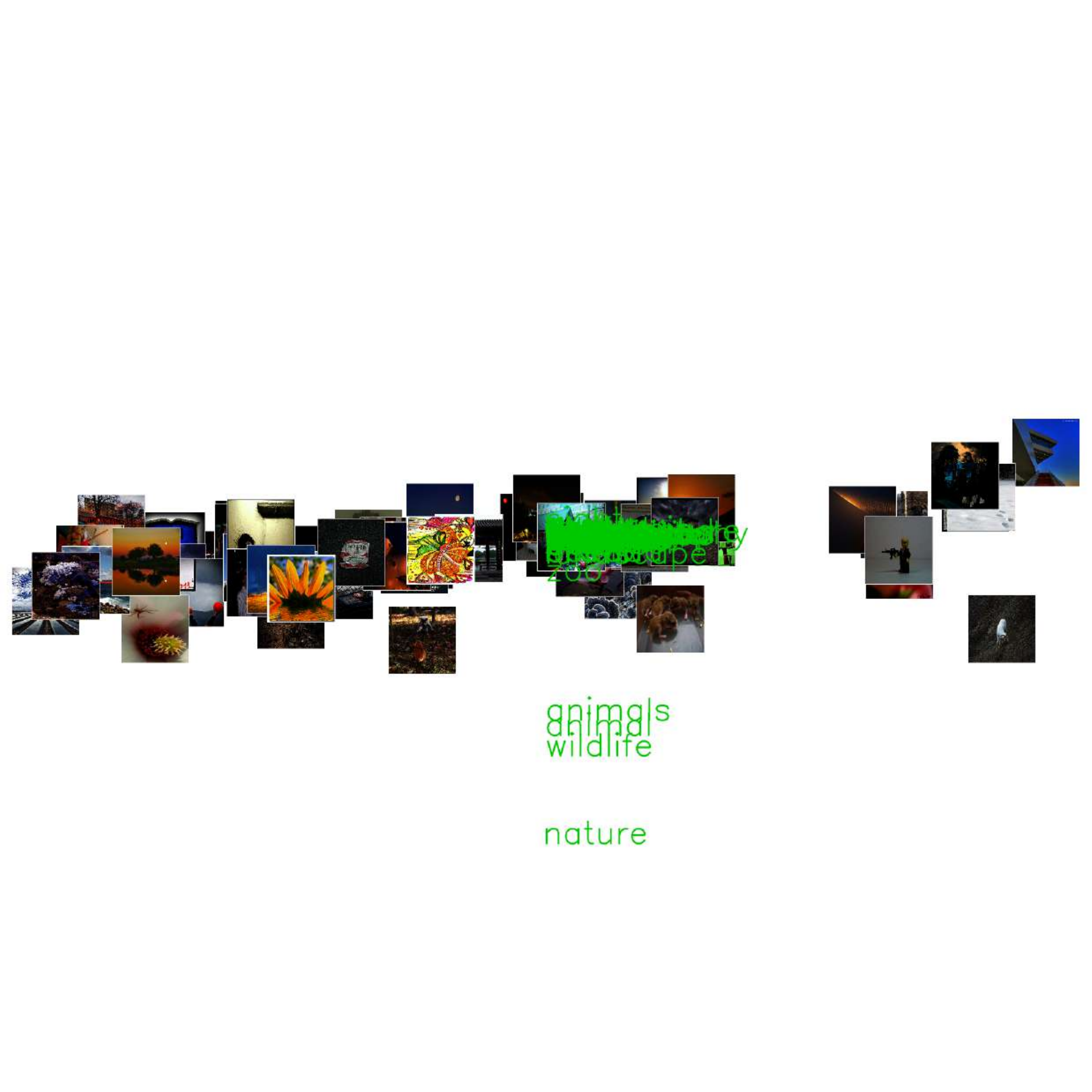}
	\label{fig:Flickr_kCDMCA}
}
\hspace{2em}
\subfigure[PMvGE (ignored)]{
		\includegraphics[width=3.5cm]{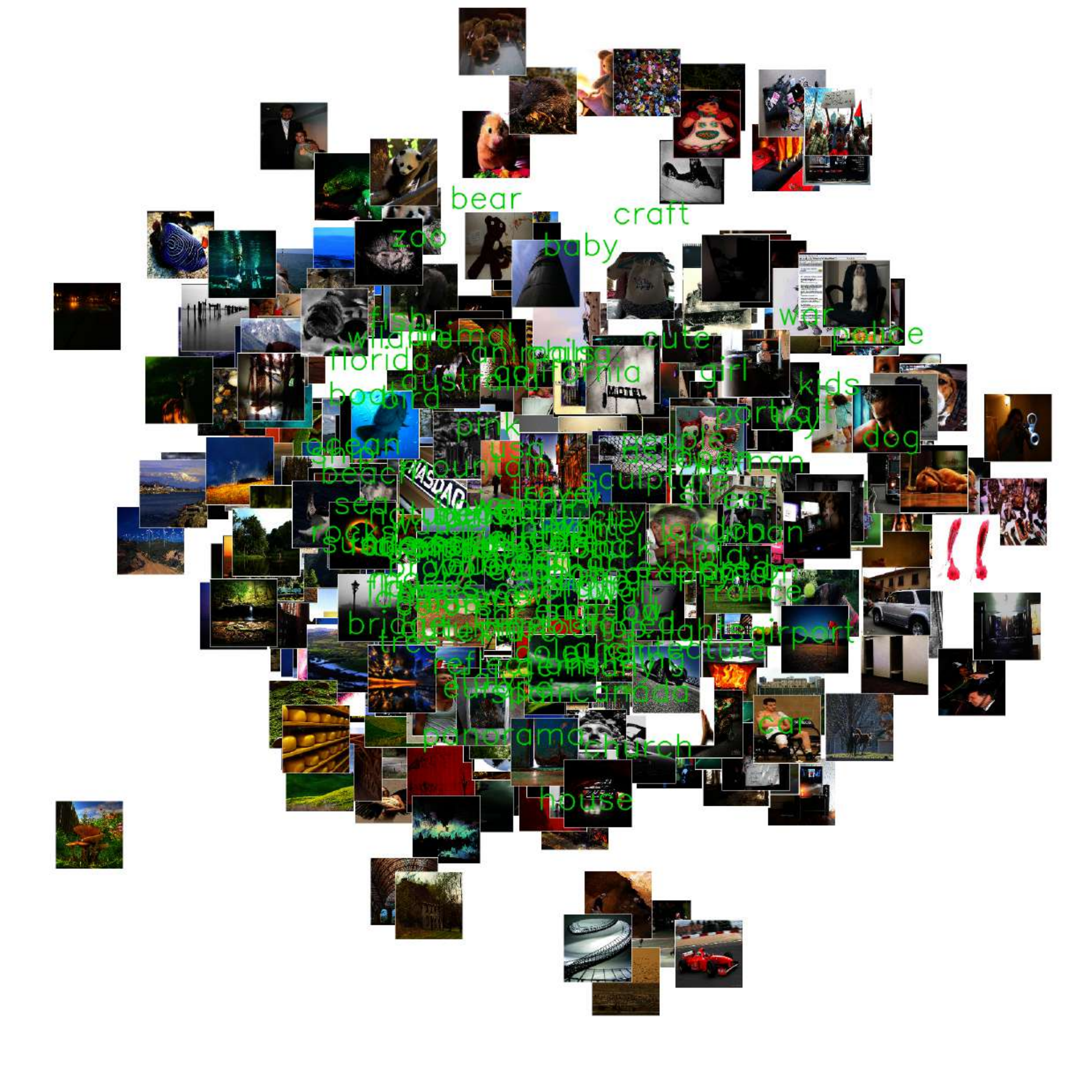}
	\label{fig:Flickr_PMvGE_ignored}
}
\centering
\subfigure[PMvGE (2nd-order)]{
		\includegraphics[width=3.5cm]{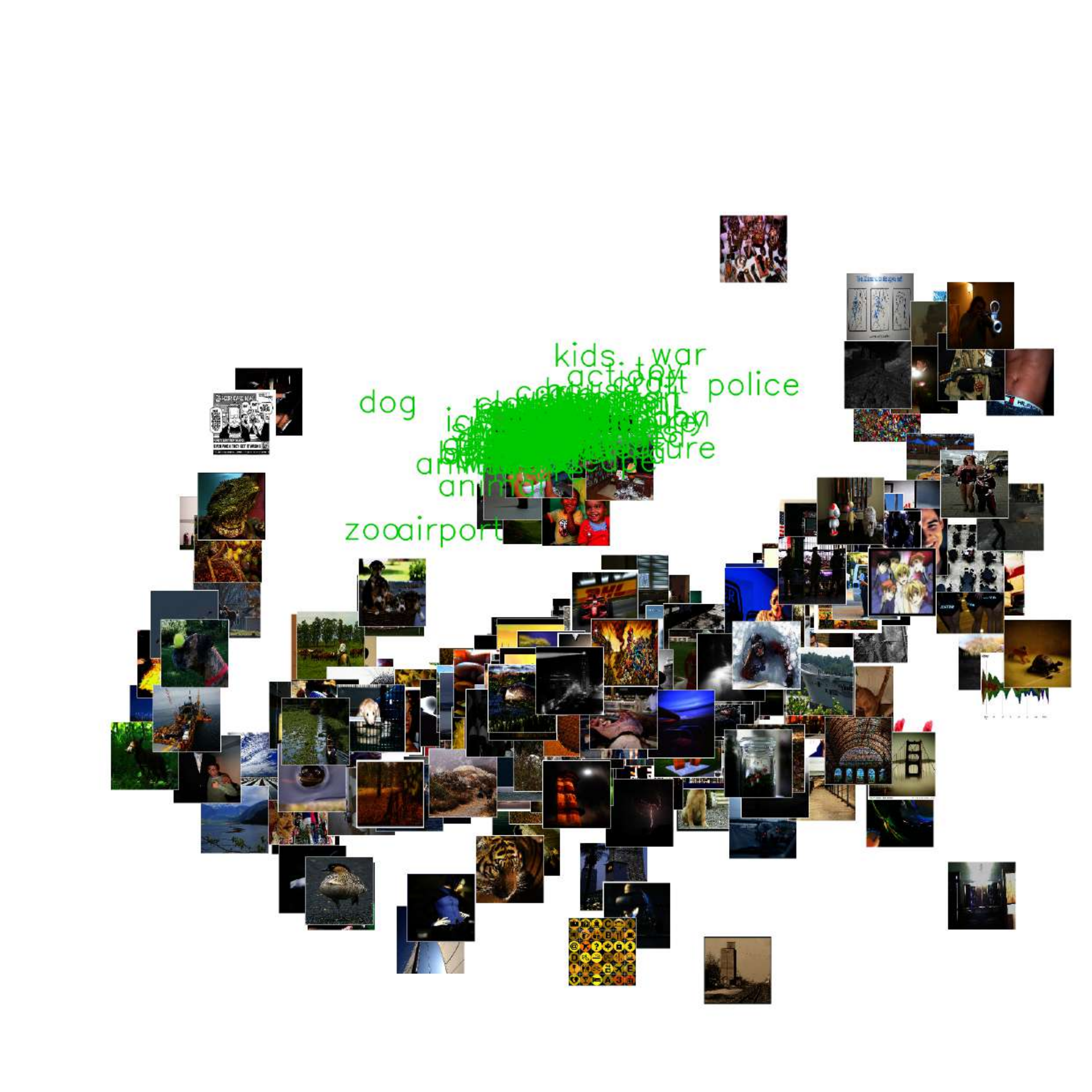}
	\label{fig:Flickr_PMvGE_winner}
}
\hspace{2em}
\subfigure[\textbf{Proposal} \protect\newline ($\beta_1=\beta_{12}=1/2,\beta_2=0)$]{
		\includegraphics[width=3.5cm]{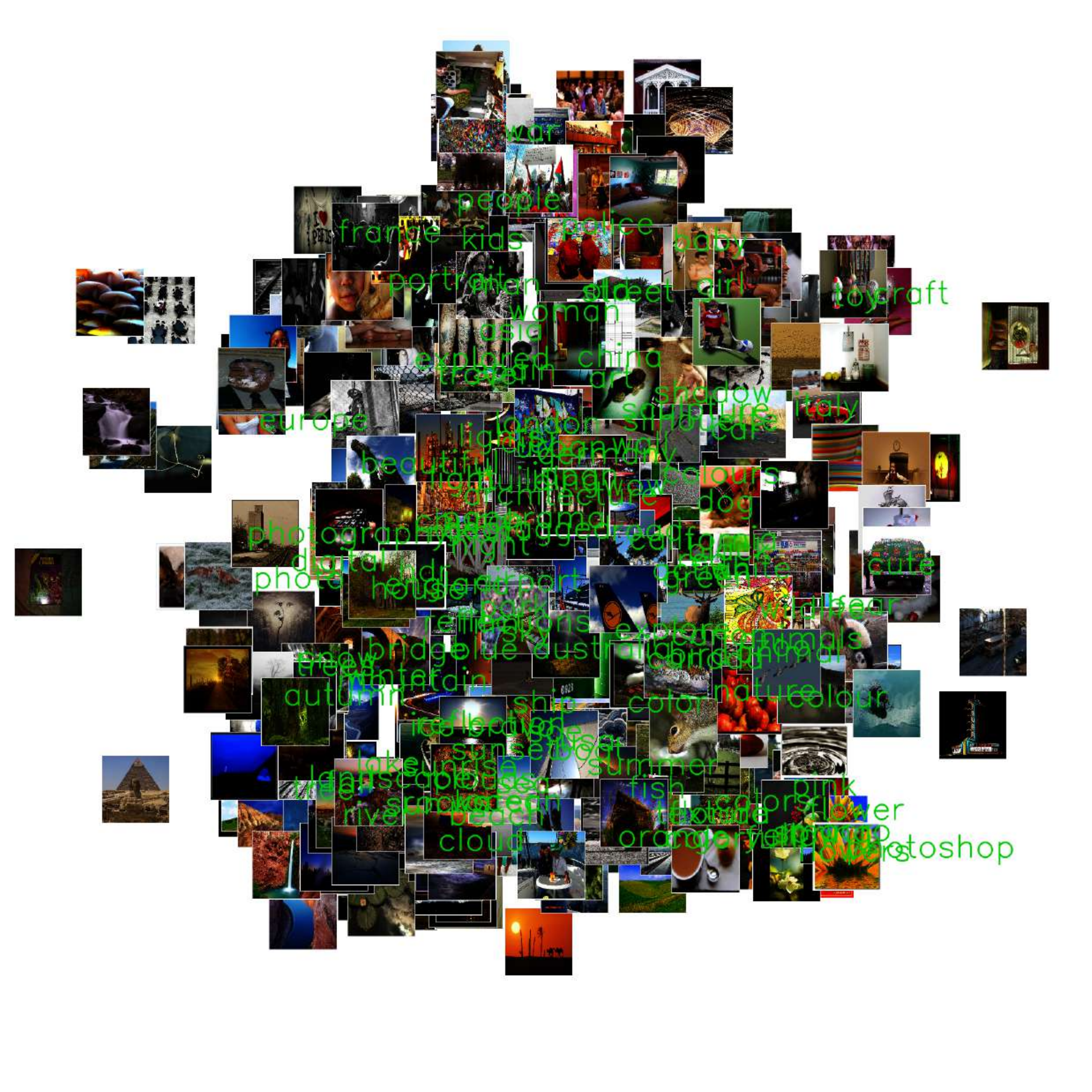}
	\label{fig:Flickr_Proposal}
}
\hspace{2em}
\subfigure[\textbf{Proposal} ($\beta_1 \propto n_1^2$, \protect\newline $\beta_{12} \propto n_1 n_2,\beta_2=0$)]{
		\includegraphics[width=3.5cm]{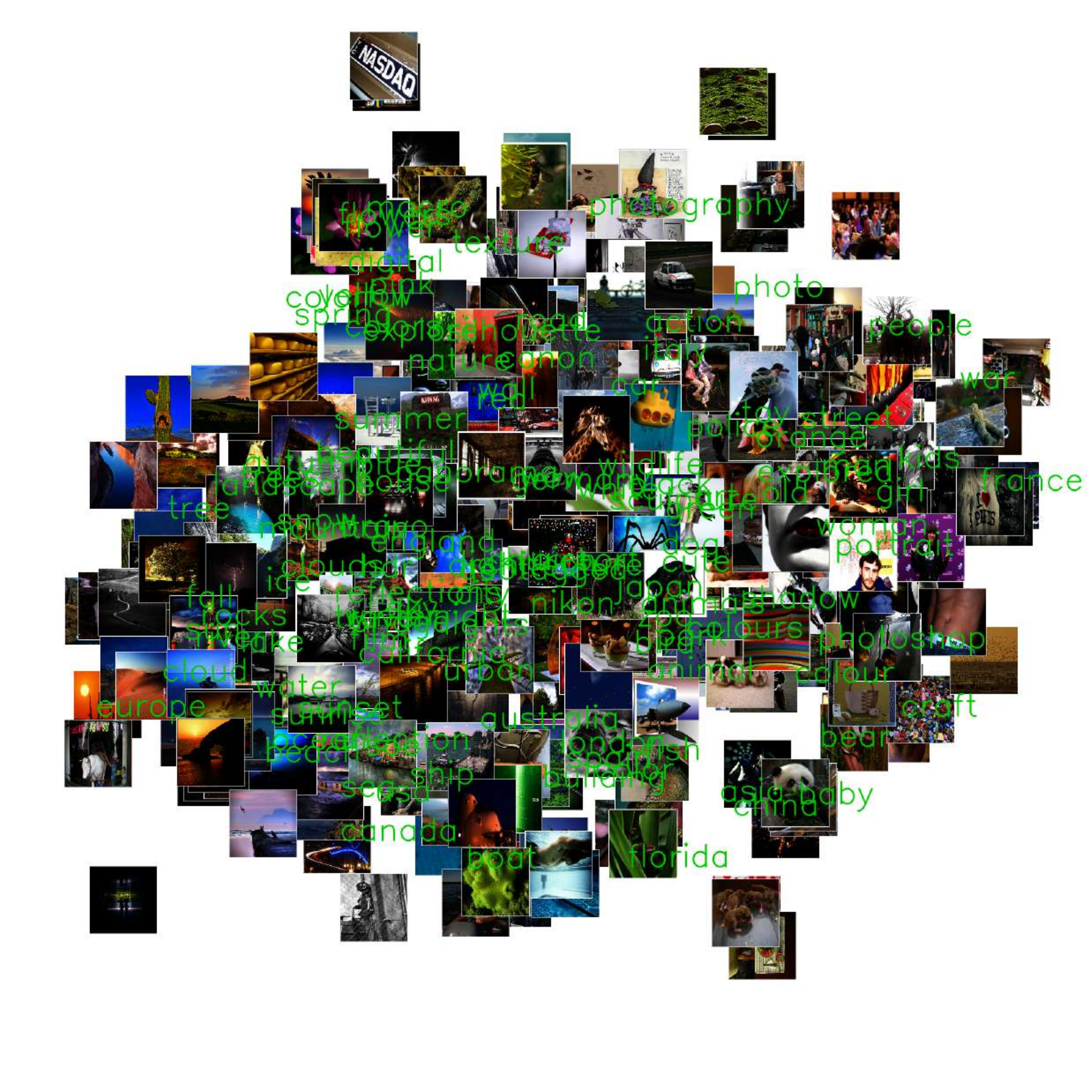}
	\label{fig:Flickr_Proposal_weight}
}
\caption{Visualization of Flickr dataset}
\label{fig:Visualization_result_Flickr}
\subfigure[PMvGE (ignored)]{
		\includegraphics[width=3.5cm]{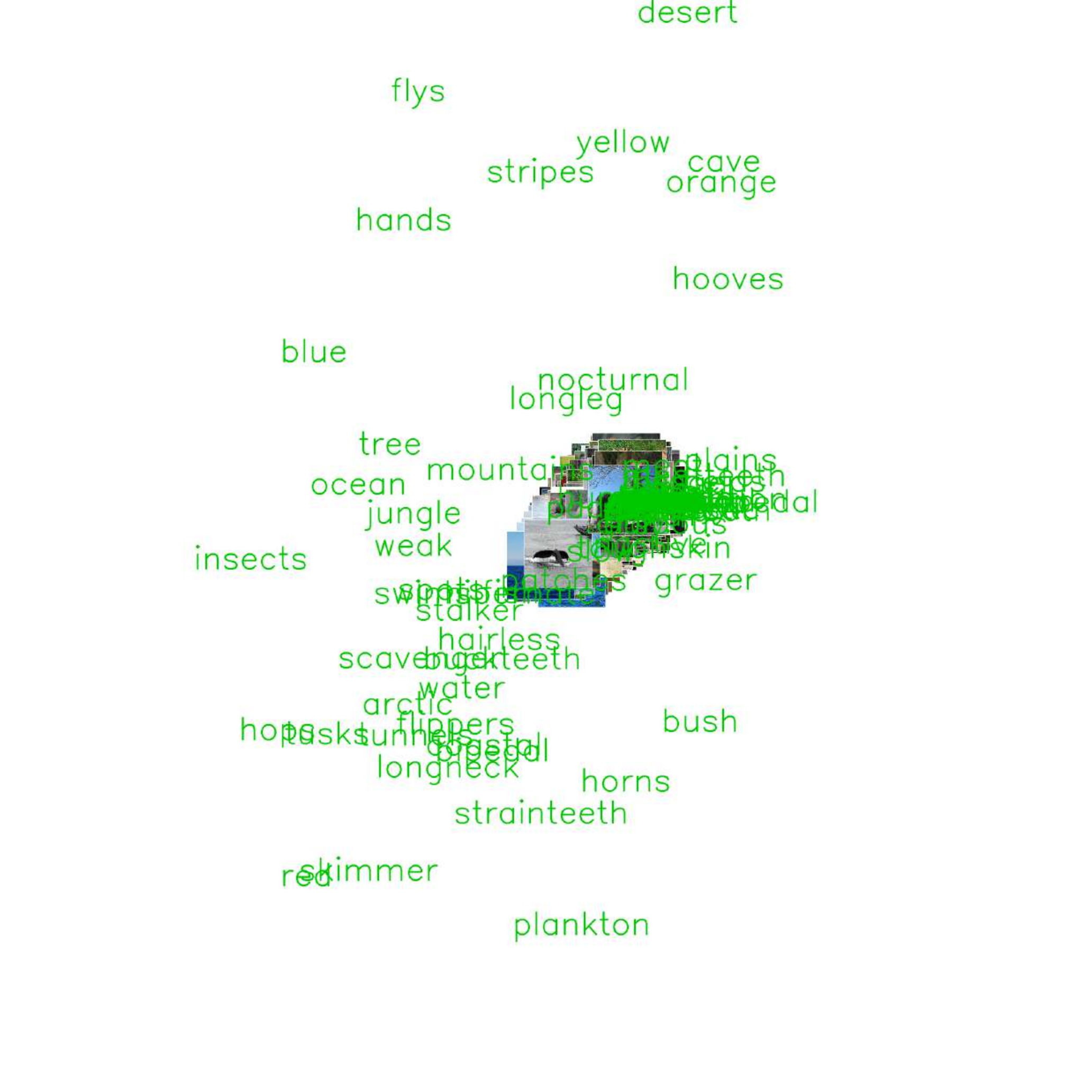}
	\label{fig:AWA2_PMvGE_ignored}
}
\hspace{2em}
\subfigure[\textbf{Proposal} \protect\newline ($\beta_1=\beta_{12}=\beta_2=1/3)$]{
		\includegraphics[width=3.5cm]{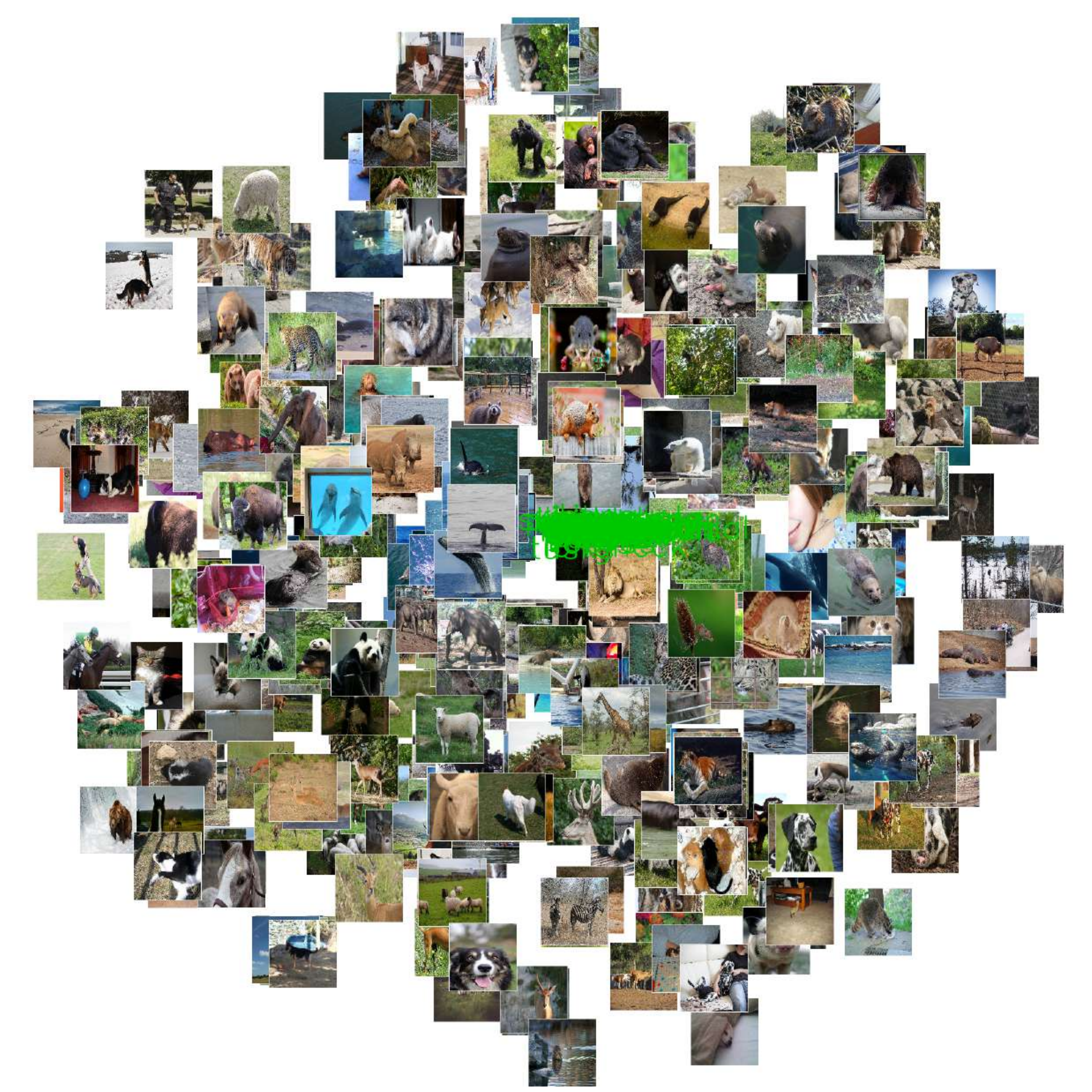}
	\label{fig:AWA2_Proposal}
}
\hspace{2em}
\subfigure[\textbf{Proposal} ($\beta_1\propto n_1^2$, \protect\newline $\beta_{12} \propto n_1 n_2, \beta_2 \propto n_2^2$)]{
		\includegraphics[width=3.5cm]{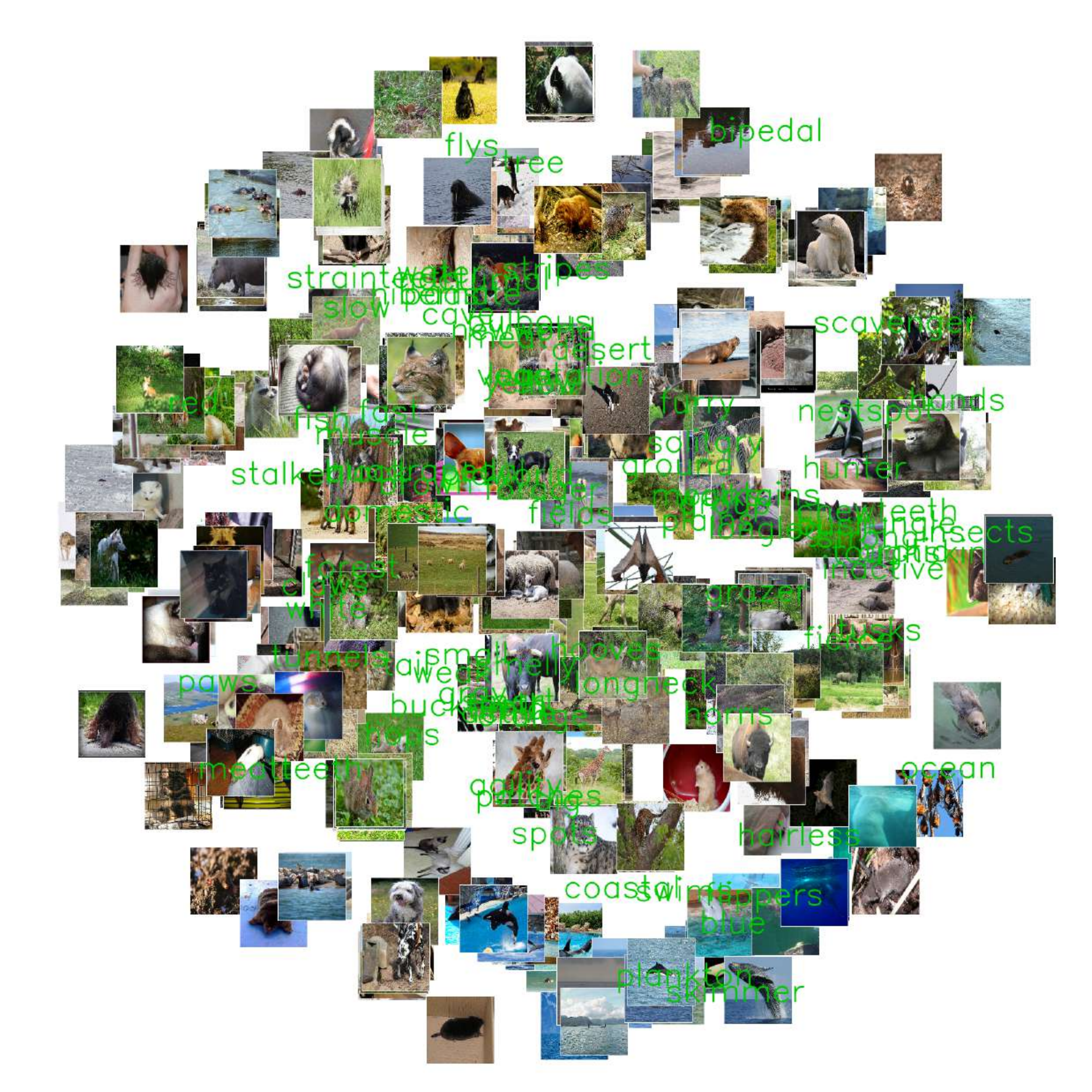}
	\label{fig:AWA2_Proposal_weight}
}
\caption{Visualization of AwA2 dataset}
\label{fig:Visualization_result_AwA}
\end{figure}

\textbf{Results:} 
(i) Feature vectors obtained in CDMCA~(Fig.~\ref{fig:Flickr_rCDMCA}) and kernel CDMCA~(Fig.~\ref{fig:Flickr_kCDMCA}) are unnaturally concentrated at the center in Flickr experiment. 
(ii) PMvGE that does not leverage within-domain graphs, i.e., PMvGE using only the across-domains graph, provides appropriate visualization for Flickr dataset~(Fig.~\ref{fig:Flickr_PMvGE_ignored}), but not for AwA2 dataset~(Fig.~\ref{fig:AWA2_PMvGE_ignored}). 
(iii) Text vectors computed via PMvGE with 2nd-order proximity are improperly shrinked for Flickr visualization~(Fig.~\ref{fig:Flickr_PMvGE_winner}). 
(iv) The proposed MR-SNE provides good visualization for both Flickr and AwA2 datasets~(Fig.~\ref{fig:Flickr_Proposal}, \ref{fig:Flickr_Proposal_weight}, \ref{fig:AWA2_Proposal}, \ref{fig:AWA2_Proposal_weight}); See Figure~\ref{fig:visualization_AWA2} for more detailed visualization. 
Above methods including the proposed MR-SNE are numerically evaluated in the following Section~\ref{subsec:experiment2} and \ref{subsec:experiment3}.

\subsection{Experiment 2: Graph Reconstruction}
\label{subsec:experiment2}

Using Flickr and AwA2 dataset, 
the proposed MR-SNE and other baselines are evaluated by ROC-AUC score for graph reconstruction, that is defined as follows: 
(i) for each query image, nearest $k$ images or text tags are predicted to be linked, 
(ii) ROC-AUC score~\citep{ROC-AUC} with respect to $k=1,2,\ldots,n_1+n_2-1$ is subsequently computed for
Flickr~($k=1,2,\ldots,1612$) and AwA2~($k=1,2,\ldots,2584$).
Therefore, this experiment considers reconstructing both within-domain and across-domains graph, though only the across-domains graph can be leveraged for training. 
Higher ROC-AUC score is better; scores for Flickr and AwA2 datasets are listed in the following Table~\ref{tab:ROC-AUC-Flickr} and \ref{tab:ROC-AUC-AwA2}, respectively. 
ROC curves for these experiments are also shown in Supplement~\ref{supp:roc}.

\begin{table}[htbp]
   \centering
   \caption{ROC-AUC score for graph reconstruction. For $K>2$, $t$-SNE is applied to obtain $2$-dim. vectors. Best score is \best{bolded} and the second best is \second{underlined}.}
\label{tab:ROC-AUC}
   \subfigure[Flickr dataset]{
   \label{tab:ROC-AUC-Flickr}
    \begin{tabular}{llccc}
         & & ignored / unnorm & 2nd-order / norm & PMI \\
    \midrule
        \multirow{3}{*}{CDMCA} & $(K=2)$ & 0.5022 & / &  / \\
        & $(K=200)$ & 0.6112& / & / \\
        & $(K=613)$ & 0.5707 & / & / \\
        \cmidrule(lr){1-1}
        \multirow{3}{*}{K-CDMCA} & $(K=2)$ & 0.4902 & / & / \\
        & $(K=200)$ & 0.5428 & / & / \\
        & $(K=613)$ & 0.5445 & / & / \\
        \cmidrule(lr){1-1}
        \multirow{2}{*}{PMvGE} & ($K=2$) & \second{0.7482} & \best{0.7918} & / \\
        & $(K=100)$ & 0.7111 & 0.7778 & / \\
        \cmidrule(lr){1-1}
        \multirow{2}{*}{\textbf{MR-SNE}} & ($\beta_1=\beta_{12}=1,\beta_2=0$) & 0.7134 & 0.7038 & 0.6690 \\
        & (\scalebox{0.8}{$\beta_1=\frac{n_1^2}{n_1^2+n_1n_2},\beta_2=0,\beta_{12}=\frac{n_1n_2}{n_1^2+n_1n_2}$}) & 0.7000 & 0.6902 & 0.6542 \\
    \end{tabular}
    }\\
   \subfigure[AwA2 dataset]{
   \label{tab:ROC-AUC-AwA2}
    \begin{tabular}{llccc}
         & & ignored / unnorm & 2nd-order / norm & PMI \\
    \midrule
        \multirow{3}{*}{CDMCA} & $(K=2)$ & 0.5650 & / &  / \\
        & $(K=200)$ & 0.6506 & / & / \\
        & $(K=613)$ & 0.6439 & / & / \\
        \cmidrule(lr){1-1}
        \multirow{3}{*}{K-CDMCA} & $(K=2)$ & 0.6427 & / & / \\
        & $(K=200)$ & 0.7283 & / & / \\
        & $(K=613)$ & 0.7285 & / & / \\
        \cmidrule(lr){1-1}
        \multirow{2}{*}{PMvGE} & ($K=2$) & 0.6576 & 0.3517 & / \\
        & $(K=100)$ & 0.3682 & 0.3657 & / \\
        \cmidrule(lr){1-1}
        \multirow{2}{*}{\textbf{MR-SNE}} & ($\beta_1=\beta_2=\beta_{12}=1$) & 0.7117 & 0.7161 & 0.7109 \\
        & (\scalebox{0.8}{$\beta_d=\frac{n_d^2}{n_1^2+n_2^2+n_1n_2},\beta_{12}=\frac{n_1n_2}{n_1^2+n_2^2+n_1n_2}$}) & 0.8033 & \second{0.8124} & \best{0.8134} \\
    \end{tabular}
    }
\end{table}

\textbf{Interpretation:} 
Overall, the proposed MR-SNE outperforms the existing CDMCA and Kernel-CDMCA. So we here mainly consider comparing the proposed MR-SNE to existing highly non-linear PMvGE. 
Firstly, Table~\ref{tab:ROC-AUC-Flickr} indicates that PMvGE demonstrates higher ROC-AUC score than others for Flickr dataset; the proposed MR-SNE is next to PMvGE. 
Unlike the Flickr dataset consisting of $n_1=1000$ images and $n_2=613$ text tags, 
AwA2 dataset contains $n_1=2500$ images and much fewer $n_2=85$ text tags; the numbers are disproportionate. 
In the AwA2 dataset, PMvGE demonstrates the worst performance whereas the proposed MR-SNE demonstrates the highest score therein. 
Thus the proposed MR-SNE is only a method that demonstrates promising performance for both of these two datasets.

\subsection{Experiment 3: Variance of Feature Vectors for Different Domains}
\label{subsec:experiment3}

Whereas ROC-AUC score for graph reconstruction is considered in the previous Section~\ref{subsec:experiment2}, 
we hereinafter examine whether the obtained feature vectors for both domains are not shrinked improperly, by computing their variance. 
In this experiment, we first compute the sample variance-covariance matrices $\hat{\Sigma}^{(\text{image})},\hat{\Sigma}^{(\text{text})} \in \mathbb{R}^{2 \times 2}$ for sets of data vectors $\mathcal{Y}^{(\text{image})}:=\{y_i^{(\text{image})}\}_{i=1}^{n_1},\mathcal{Y}^{(\text{text})}:=\{y_j^{(\text{text})}\}_{j=1}^{n_2}$, respectively. 
Subsequently, $r:=\frac{\text{tr}\hat{\Sigma}^{(\text{image})}}{\text{tr}\hat{\Sigma}^{(\text{text})}}$ is computed for each of datasets in Table~\ref{tab:experiment_3} below, where $\text{tr}A=\sum_{i=1}^{d}a_{ii}$ denotes trace of $A=(a_{ij})$.

\begin{table}[htbp]
    \centering
   \subfigure[Flickr dataset]{
   \label{tab:Variance-Flickr}
    \begin{tabular}{llccc}
         & &  ignored / unnorm & 2nd-order / norm & PMI \\
        \hline
        \multirow{2}{*}{PMvGE} & $(K=2)$ & 1.448 & 1.666 & / \\
         & $(K=100)$ & 1.192 & 0.719 & / \\
        \cmidrule(lr){1-1}
        \multirow{2}{*}{\textbf{MR-SNE}} & ($\beta_1=\beta_{12}=1,\beta_2=0$) & 1.706 & 1.388 & \second{1.151} \\
         & (\scalebox{0.8}{$\beta_1=\frac{n_1^2}{n_1^2+n_1n_2},\beta_2=0,\beta_{12}=\frac{n_1n_2}{n_1^2+n_1n_2}$}) & 1.556 & 1.475 & \best{1.067} \\
    \end{tabular}
    }
    \subfigure[AwA2 dataset]{
    \label{tab:Variance-AwA2}
    \begin{tabular}{llccc}
         & &  ignored / unnorm & 2nd-order / norm & PMI \\
        \hline
        \multirow{2}{*}{PMvGE} & $(K=2)$ & 0.031 & 0.001 & / \\
        & $(K=100)$ & 0.170 & 0.282 & / \\
        \cmidrule(lr){1-1}
        \multirow{2}{*}{\textbf{MR-SNE}} & ($\beta_1=\beta_2=\beta_{12}=1$) & 56.563 & 26.139 & 27.800 \\
         & (\scalebox{0.8}{$\beta_d=\frac{n_d^2}{n_1^2+n_2^2+n_1n_2},\beta_{12}=\frac{n_1n_2}{n_1^2+n_2^2+n_1n_2}$}) & \best{1.087} & 1.190 & \second{1.165} \\
    \end{tabular}
    }
\caption{$r:=\frac{\text{tr}\hat{\Sigma}^{(\text{image})}}{\text{tr}\hat{\Sigma}^{(\text{text})}}$ is listed; $t$-SNE is applied to obtain $2$-dimensional vectors for $K>2$.}
\label{tab:experiment_3}
\end{table}

\textbf{Interpretation:} 
Roughly speaking, the vectors are improperly shrinked if $r$ is far away from the value $1$. 
For instance, PMvGE~($K=2$) using $2$nd-order proximity leads to $r=0.001$, indicating that $\text{tr}\hat{\Sigma}^{(\text{image})}$ is much smaller than $\text{tr}\hat{\Sigma}^{(\text{text})}$. Image vectors are then obviously improperly shrinked. 
On the other hand, MR-SNE with the adaptive weights satisfies $r \approx 1$ for both of datasets; MR-SNE is more stable than the existing PMvGE.

\subsection{Additional Numerical Experiments}
Additional numerical experiments are also conducted in Supplement~\ref{supp:additional_experiments}. 
Overall, the proposed MR-SNE demonstrates better score than other baselines for $K=2$; thus MR-SNE is expected to provide better visualization. 
Also see Supplement~\ref{supp:stochastic_pmvge} for PMvGE equipped with stochastic neighbor graph.

\section{Conclusion and Future Works}
\label{sec:conclusion}
We proposed MR-SNE, that visualizes multimodal relational data, such as images, texts, and their relations in Flickr~\citep{Flickr} and AwA2~\citep{AWA2} datasets. 
MR-SNE (i) first computes augmented relations for the multimodal relational data, where the relations across domains are observed and those within domains are computed via the observed data vectors, and (2) jointly embeds the multimodal relational data to a common low-dimensional subspace. 
Numerical experiments were conducted for performing MR-SNE and other alternatives for visualizing multimodal relational data. 
For future work, it would be worthwhile to (i) consider more efficient computation of MR-SNE by referring to a multi-core extension of $t$-SNE~\citep{van2014accelerating,multicore-tsne}, and (ii) develop more effective metric for evaluating visualization methods.

\clearpage

\bibliographystyle{apalike}

\clearpage

\appendix

\begin{flushleft}
\textbf{\Large Supplementary Material:} \par
{\Large Stochastic Neighbor Embedding of Multimodal Relational Data for Image-Text Simultaneous Visualization}
\end{flushleft}

\section{Algorithm}
\label{supp:algorithm}

The following Algorithm~\ref{alg:proposal} describes the optimization procedure of the proposed MR-SNE, that is also explained in Section~\ref{subsec:proposal}.

 \begin{algorithm}[htbp]
 \caption{Multimodal Relational Stochastic Neighbor Embedding ($D=2$).}
 \label{alg:proposal}
 \begin{algorithmic}
 \renewcommand{\algorithmicrequire}{\textbf{Input:}}
 \renewcommand{\algorithmicensure}{\textbf{Output:}}
 \REQUIRE Data vectors $\{ x^{(1)}_i\} ^{n_1}_{i=1}\in\mathbb{R}^{d_1}$, $\{ x^{(2)}_j\} ^{n_2}_{j=1} \in \mathbb{R}^{d_2}$, across-domains graph $W = (w_{ij}) \in \mathbb{R}^{n_1 \times n_2}$, and dimensionality $K$ of the common subspace.\\
 For cost function: perplexity $\lambda>0$, weights $\beta_1,\beta_2,\beta_{12} \ge 0$ satisfying $\beta_1+\beta_2+\beta_{12}=1$. \\
 For optimization: number of iterations $T \in \mathbb{N}$, initial learning rate $\eta>0$, momentum $\alpha>0$. 
 
 \ENSURE Low-dimensional vectors $\{y^{(1)}_i\}_{i=1}^{n_1},\{y^{(2)}_j\}_{j=1}^{n_2} \subset \mathbb{R}^K$.
 
 \STATE Parameters $\{\sigma_i\}_{i=1}^{n}$ are computed via the user-specified perplexity. 
 
 \STATE Compute stochastic neighbor graphs $P^{(1)},P^{(2)}$ and the normalized across-domains graph $R$ by the step~(i) in Section~\ref{subsec:proposal}; 
 subsequently construct the matrix $\tilde{P}$ from the matrices $P^{(1)},P^{(2)},R$. 
 
 \STATE Randomly initialize the entries in $Y=(y^{(1)}_1,y^{(1)}_2,\ldots,y^{(1)}_{n_1},y^{(2)}_1,y^{(2)}_2,\ldots,y^{(2)}_{n_2})$ with the normal distribution $N(0,10^{-4})$, and set $Y^{(0)}=Y^{(1)}$. 
 \FOR {$t = 1$ to $T$}
  \STATE Compute $\tilde{Q}$ by the step~(ii) in Section~\ref{subsec:proposal}. 
  \STATE Update $Y^{(t)}$ by leveraging the full-batch gradient descent with momentum: 
  \begin{align}
  Y^{(t+1)}= Y^{(t)}-\eta_t \frac{\partial \tilde{C}}{\partial Y^{(t)}}+ \alpha_t(Y^{(t)}-Y^{(t-1)}),
  \end{align}
  where $\tilde{C}$ is defined in eq.~(\ref{eq:objective_proposal}).
  \STATE Multiply $1/10$ to the learning rate for each $400$ iterations: $\eta_{t+1}=\eta_t/10$ for $t=400k \: (k \in \mathbb{N})$ and $\eta_{t+1}=\eta_t$ otherwise. The momentum is kept fixed as $\alpha_t = \alpha$ in our implementation.
  \ENDFOR
  \STATE Output $Y^{(T+1)}$.
 \end{algorithmic} 
 \end{algorithm}

\subsection{Computational time}
\label{supp:computational_time}

Numerical experiments were performed on a server with
 (CPU)~3.70 GHz Intel Xeon E5-1620 v2 
 and (GPU)~NVIDIA GP102 TITAN Xp.
The computation time was roughly $8$ hours and $11.5$ hours for training MR-SNE on the Flickr and AwA2 datasets, respectively. 
In this this study, we did not attempt an efficient implementation of MR-SNE,
and it would be a future work to consider more efficient computation by referring to a multi-core extension of $t$-SNE~\citep{van2014accelerating,multicore-tsne}.

\clearpage
\section{Visualization of Flickr dataset}
\label{supp:visualization_Flickr}

\begin{figure}[htbp]
\centering
\includegraphics[width=0.80\textwidth]{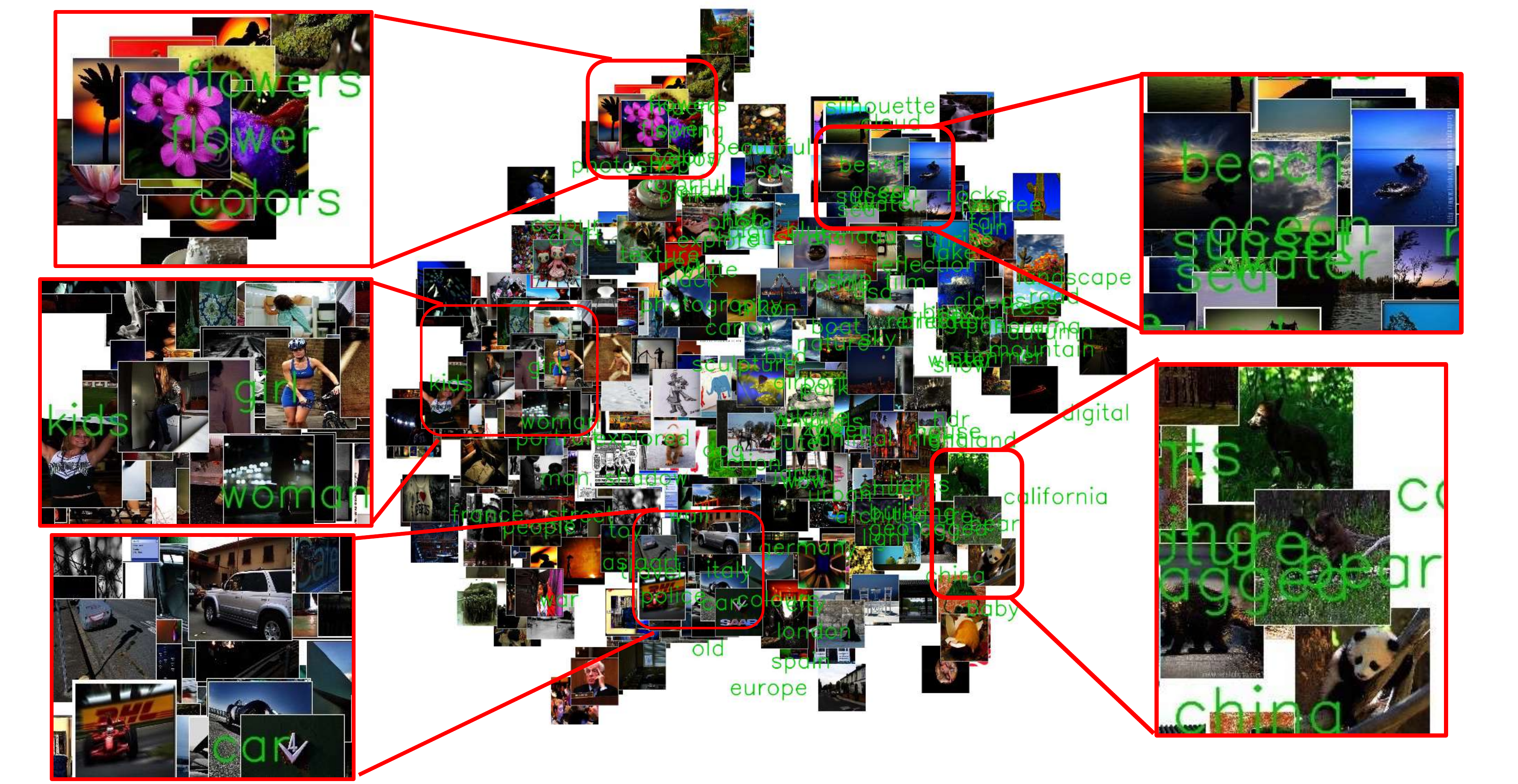}
\caption{Simultaneous visualization of $n_1=1000$ images and most frequently appeared $n_2=613$ text tags in Flickr dataset.
Randomly selected 500 images and the 120 most frequently used tags are displayed in green.
These vectors are embedded in $K=2$ dimensional common subspace, by applying the proposed MR-SNE. }
\label{fig:visualization_Flickr}
\end{figure}

Similarly to Figure~\ref{fig:visualization_AWA2} of AwA2 dataset, 
visualization of Flickr dataset is shown in Figure~\ref{fig:visualization_Flickr}.

\section{ROC-Curve}
\label{supp:roc}

\begin{figure}[htbp]
\centering
\subfigure[Flickr]{
		\includegraphics[height=3.0cm]{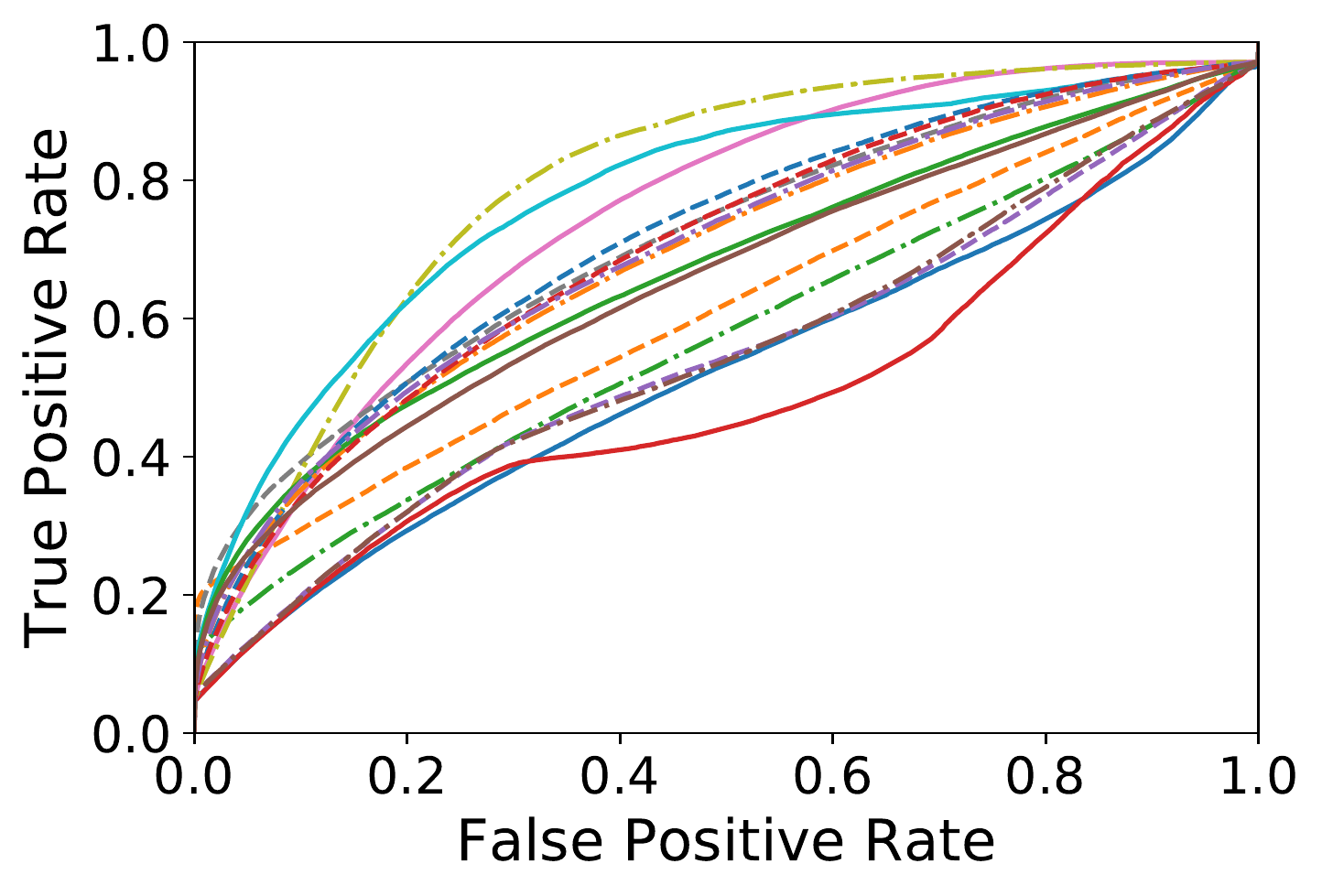}
	\label{fig:ROC-curve_Flickr}
}
\subfigure[AwA2]{
		\includegraphics[height=3.0cm]{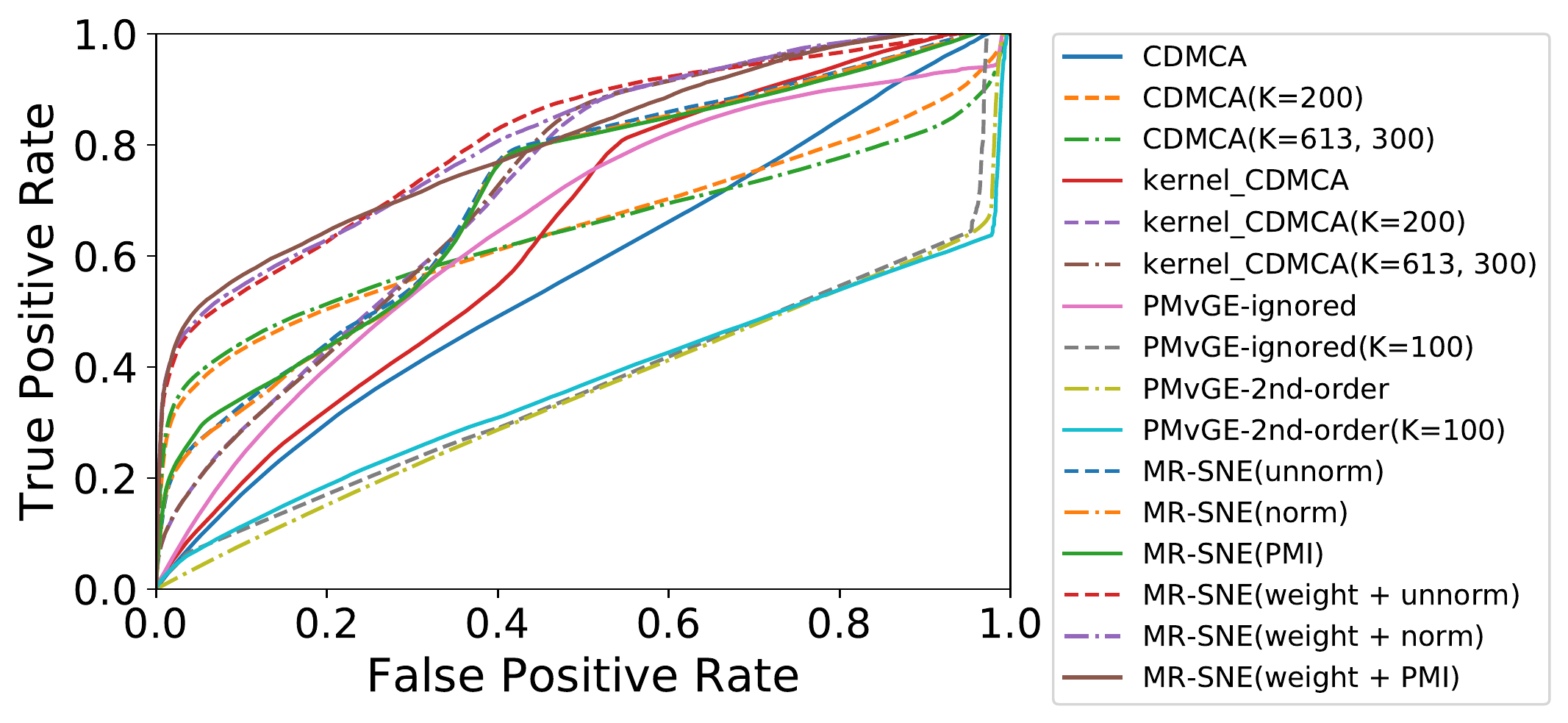}
	\label{fig:ROC-curve_AwA2}
}
\caption{ROC-curve}
\label{fig:ROC curve}
\end{figure}

In Experiment~2 shown in Section~\ref{subsec:experiment2}, proposed MR-SNE is compared to the baselines by ROC-AUC score for the task of graph reconstruction. 
ROC curve is shown in Figure~\ref{fig:ROC curve}, for each of Flickr and AwA2 datasets.
In these plots,  ``MR-SNE (weight + ...)'' represents that the weights $\beta_1,\beta_2,\beta_{12}$ therein are specified by  $\beta_1=\frac{n_1^2}{n_1^2+n_1n_2},\beta_2=0,\beta_{12}=\frac{n_1n_2}{n_1^2+n_1n_2}$ for Flickr dataset and $\beta_d=\frac{n_d^2}{n_1^2+n_2^2+n_1n_2},\beta_{12}=\frac{n_1n_2}{n_1^2+n_2^2+n_1n_2}$ for AwA2 dataset. 

\clearpage
\section{Additional Numerical Experiments}
\label{supp:additional_experiments}

\begin{figure}[htbp]
\centering
\includegraphics[width=0.70\textwidth]{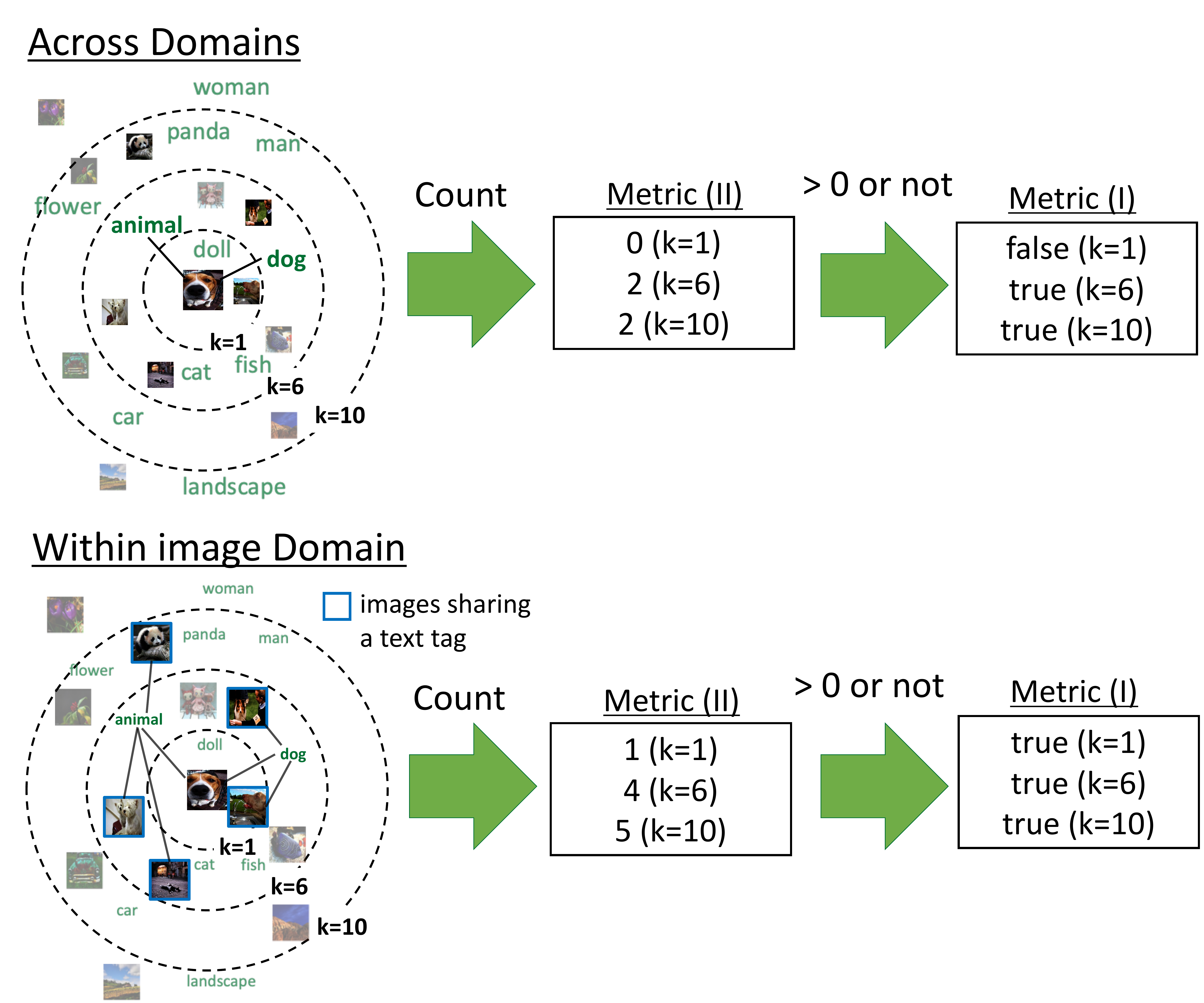}
\caption{Four metrics used in supplementary numerical experiments. For a image query (dog), we compute the scores: the scores are averaged over all the queries.}
\label{fig:evaluation}
\end{figure}

In addition to Experiment~1--3 shown in Section~\ref{subsec:experiment1}--\ref{subsec:experiment3}, the proposed MR-SNE and other baselines are here evaluated by the following four evaluation metrics, that are also shown in Figure~\ref{fig:evaluation}. 
Amongst all the evaluations, $k$-nearest neighbors are computed via the obtained feature vectors in the common subspace.

\begin{enumerate}[{(1)}]
\item \textbf{Metric (I) across domains}: 
For each query image, we examine whether at least one associated text tag is in its $k$-nearest neighbor text tags (searched over all the text tags); we then compute the proportion of such queries to all the query images. 

\item \textbf{Metric (I) within image domain}: 
For each query image, we examine whether at least one image in its $k$-nearest neighbor (searched over all the images) shares at least one text tag with the query; we then compute the proportion of such queries to all the query images. 

\item \textbf{Metric (II) across domains}: 
For each query image, we compute the number of associated text tags in its $k$-nearest neighbor text tags (searched over all the text tags); the numbers are also averaged over all the query images.

\item \textbf{Metric (II) within image domain}: 
For each query image, we compute the number of images sharing at least one text tag with the query, in its $k$-nearest neighbor images (searched over all the images); the numbers are also averaged over all the query images. 
\end{enumerate}

Higher score is better; scores for Flickr and AwA2 datasets are listed in the following Table~\ref{tab:experiment_d1}\subref{table:Accuracy-across-domains-Flickr}--\ref{tab:experiment_d4}\subref{table:Number-within-domain-AwA2}. In these tables, ``MR-SNE (weight + ...)'' represents that $\beta_1,\beta_2,\beta_{12}$ are specified as defined in Supplement~\ref{supp:roc}.

\clearpage
\subsection{Metric (I) across domains}
\label{subsec:metric1_across}

\begin{table}[htbp]
\centering
\caption{Evaluated by metric (I) across domains. For $K>2$, $t$-SNE is applied to obtain $2$-dim. vectors. Best score is \best{bolded} and the second best is \second{underlined}.}
\label{tab:experiment_d1}
 \subfigure[Flickr dataset]{
 \label{table:Accuracy-across-domains-Flickr}
 \centering
 \scalebox{0.9}{
  \begin{tabular}{llccccccccc}
   \hline
   &  & $k=$1 & 2 & 5 & 10 & 20 & 50 & 100 & 200 & 500 \\
   \hline
   \multirow{3}{*}{CDMCA} & ($K=2$) & 0.021 & 0.042 & 0.085 & 0.141 & 0.221 & 0.382 & 0.531 & 0.707 & 0.912 \\
    & ($K=200$) & \second{0.758} & \second{0.808} & \best{0.843} & \second{0.852} & 0.871 & 0.891 & 0.909 & 0.923 & 0.949 \\
   & ($K=613$) & \best{0.806} & \best{0.821} & \second{0.838} & \best{0.856} & 0.884 & \second{0.929} & \best{0.944} & \best{0.952} & \best{0.955} \\
   \cmidrule(lr){1-1}
   \multirow{3}{*}{K-CDMCA} & ($K=2$) & 0.024 & 0.030 & 0.062 & 0.121 & 0.229 & 0.379 & 0.517 & 0.665 & 0.888 \\
   & ($K=200$) & 0.046 & 0.056 & 0.103 & 0.143 & 0209 & 0.365 & 0.503 & 0.706 & 0.912 \\
   & ($K=613$) & 0.059 & 0.064 & 0.102 & 0.166 & 0.253 & 0.384 & 0.523 & 0.684 & 0.917 \\
   \cmidrule(lr){1-1}
   \multirow{2}{*}{PMvGE-ignored} & ($K=2$) & 0.078 & 0.136 & 0.263 & 0.414 & 0.629 & 0.837 & 0.928 & 0.952 & \best{0.955} \\
    & ($K=100$) & 0.544 & 0.651 & 0.763 & 0.824 & 0.871 & 0.908 & 0.940 & \best{0.952} & \best{0.955} \\
   \cmidrule(lr){1-1}
   \multirow{2}{*}{PMvGE-2nd-order} & ($K=2$) & 0.015 & 0.074 & 0.196 & 0.371 & 0.569 & 0.805 & 0.895 & 0.943 & 0.955 \\
    & ($K=100$) & 0.035 & 0.040 & 0.050 & 0.075 & 0.127 & 0.210 & 0.315 & 0.538 & 0.920 \\
    \cmidrule(lr){1-1}
   \multirow{6}{*}{ \textbf{MR-SNE}}
   & \textbf{(unnorm)} & 0.163 & 0.280 & 0.464 & 0.632 & 0.774 & 0.870 & 0.904 & 0.930 & 0.952 \\
   & \textbf{(norm)} & 0.234 & 0.363 & 0.578 & 0.733 & 0.848 & 0.926 & 0.942 & \second{0.949} & \best{0.955} \\
   & \textbf{(PMI)} & 0.319 & 0.485 & 0.729 & 0.844 & \best{0.905} & \best{0.938} & \second{0.943} & \best{0.952} & \best{0.955} \\
   & \textbf{(weight+unnorm)} & 0.163 & 0.258 & 0.458 & 0.621 & 0.741 & 0.848 & 0.889 & 0.925 & 0.950 \\
   & \textbf{(weight+norm)} & 0.208 & 0.349 & 0.568 & 0.725 & 0.829 & 0.902 & 0.930 & 0.946 & \second{0.954} \\
   & \textbf{(weight+PMI)} & 0.313 & 0.462 & 0.674 & 0.809 & \second{0.889} & 0.927 & 0.942 & 0.948 & \second{0.954} \\
   \hline
  \end{tabular}
  }
}
\subfigure[AwA2 dataset]{
 \label{table:Accuracy-across-domains-AwA2}
 \centering
 \scalebox{0.9}{
  \begin{tabular}{llcccccccc}
   \hline
   &  & $k=$1 & 2 & 3 & 5 & 10 & 20 & 30 & 50 \\
   \hline
   \multirow{3}{*}{CDMCA} & ($K=2$) & 0.579 & 0.798 & 0.893 & 0.970 & 0.994 & 0.999 & \best{1.000} & \best{1.000} \\
   & ($K=200$) & 0.246 & 0.502 & 0.635 & 0.798 & 0.939 & \best{1.000} & \best{1.000} & \best{1.000}  \\
   & ($K=300$) & 0.264 & 0.539 & 0.718 & 0.858 & 0.995 & \best{1.000} & \best{1.000} & \best{1.000}  \\
   \cmidrule(lr){1-1}
   \multirow{3}{*}{K-CDMCA} & ($K=2$) & 0.576 & 0.754 & 0.843 & 0.962 & \best{1.000} & \best{1.000} & \best{1.000} & \best{1.000}  \\
   & ($K=200$) & 0.174 & 0.490 & 0.713 & 0.874 & 0.952 & 0.986 & \best{1.000} & \best{1.000}  \\
   & ($K=300$) & 0.192 & 0.497 & 0.704 & 0.855 & 0.955 & 0.986 & \best{1.000} & \best{1.000}  \\
   \cmidrule(lr){1-1}
   \multirow{2}{*}{PMvGE-ignored} & ($K=2$) & \second{0.698} & \second{0.887 }& \second{0.948} & \second{0.988} & \second{0.999} & \best{1.000} & \best{1.000} & \best{1.000} \\
   & ($K=100$) & 0.140 & 0.220 & 0.540 & 0.720 & 0.904 & 0.924 & 0.944 & \best{1.000}  \\
   \cmidrule(lr){1-1}
   \multirow{2}{*}{PMvGE-2nd-order} & ($K=2$) & 0.685 & \best{0.950} & \best{0.997} & \best{1.000} & \best{1.000} & \best{1.000} & \best{1.000} & \best{1.000}  \\
   & ($K=100$) & 0.100 & 0.180 & 0.180 & 0.800 & 0.974 & \best{1.000} & \best{1.000} & \best{1.000} \\
   \cmidrule(lr){1-1}
   \multirow{6}{*}{\textbf{MR-SNE}} 
   & \textbf{(unnorm)} & 0.539 & 0.756 & 0.804 & 0.923 & 0.984 & \second{0.999} & \best{1.000} & \best{1.000} \\
   & \textbf{(norm)} & 0.232 & 0.389 & 0.669 & 0.798 & 0.906 & 0.987 & \best{1.000} & \best{1.000} \\
   & \textbf{(PMI)} & 0.128 & 0.291 & 0.389 & 0.516 & 0.712 & 0.947 & \second{0.999} & \best{1.000} \\
   & \textbf{(weight+unnorm)} & 0.520 & 0.751 & 0.863 & 0.919 & 0.995 & \second{0.999} & \best{1.000} & \best{1.000} \\
   & \textbf{(weight+norm)} & 0.620 & 0.758 & 0.824 & 0.925 & 0.997 & \best{1.000} & \best{1.000} & \best{1.000} \\
   & \textbf{(weight+PMI)} & \best{0.703} & 0.841 & 0.881 & 0.961 & 0.995 & \second{0.999} & \best{1.000} & \best{1.000} \\
   \hline
  \end{tabular}
  }
}
\end{table}
Overall, CDMCA~($K=613$) demonstrates better scores in Flickr dataset. However, the good scores are for $K=613$, meaning that the dimension of feature vector is very high; when considering only the case $K=2$, that is usually used for visualization, MR-SNE~(PMI) demonstrates the best performance. Thus MR-SNE is expected for providing better visualization. 
For AwA2 dataset, PMI-2nd~($K=2$) demonstrates the best scores. Overall, $K=2$ is better than $K>2$ amongst all the baselines; only the low-dimension is required for representing associations across domains, for AwA2 dataset.

\clearpage
\subsection{Metric (I) within image domain}
\label{subsec:metric1_within}

\begin{table}[htbp]
\centering
\caption{Evaluated by metric (I) within image domain. For $K>2$, $t$-SNE is applied to obtain $2$-dim. vectors. Best score is \best{bolded} and the second best is \second{underlined}.}
\label{tab:experiment_d2}
 \subfigure[Flickr dataset]{
 \label{table:Accuracy-within-domain-Flickr}
 \centering
 \scalebox{0.8}{
  \begin{tabular}{llccccccccc}
   \hline
     &  & $k=$1 & 2 & 5 & 10 & 20 & 50 & 100 & 200 & 500 \\
   \hline
   \multirow{3}{*}{CDMCA} & ($K=2$) & 0.170 & 0.272 & 0.421 & 0.522 & 0.634 & 0.760 & 0.838 & 0.903 & 0.950 \\
   & ($K=200$) & \best{0.887} & \best{0.904} & \best{0.913} & \best{0.926} & \best{0.939} & 0.944 & 0.947 & 0.949 & 0.953 \\
   & ($K=613$) & \second{0.643} & 0.722 & 0.790 & 0.841 & 0.884 & 0.918 & 0.937 & 0.948 & \best{0.955} \\
   \cmidrule(lr){1-1}
   \multirow{3}{*}{K-CDMCA} & ($K=2$) & 0.094 & 0.169 & 0.329 & 0.511 & 0.655 & 0.809 & 0.876 & 0.925 & 0.949 \\
   & ($K=200$) & 0.191 & 0.280 & 0.441 & 0.566 & 0.699 & 0.809 & 0.874 & 0.915 & 0.944 \\
   & ($K=613$) & 0.194 & 0.291 & 0.434 & 0.550 & 0.677 & 0.805 & 0.870 & 0.922 & 0.946 \\
   \cmidrule(lr){1-1}
   \multirow{2}{*}{PMvGE-ignored} & ($K=2$) & 0.213 & 0.337 & 0.536 & 0.700 & 0.819 & 0.917 & 0.949 & \best{0.955} & \best{0.955} \\
   & ($K=100$) & 0.640 & 0.762 & \second{0.874} & \second{0.911} & \second{0.938} & \second{0.948} & \best{0.954} & \best{0.955} & \best{0.955} \\
   \cmidrule(lr){1-1}
   \multirow{2}{*}{PMvGE-2nd-order} & ($K=2$) & 0.210 & 0.322 & 0.529 & 0.664 & 0.787 & 0.881 & 0.929 & 0.949 & \best{0.955} \\
   & ($K=100$) & 0.641 & 0.758 & 0.852 & 0.896 & 0.922 & 0.941 & 0.946 & 0.950 & 0.953 \\
   \cmidrule(lr){1-1}
   \multirow{6}{*}{\textbf{MR-SNE}} 
   & \textbf{(unnorm)} & 0.266 & 0.409 & 0.602 & 0.731 & 0.821 & 0.897 & 0.923 & 0.939 & 0.951 \\
   & \textbf{(norm)} & 0.296 & 0.450 & 0.689 & 0.806 & 0.888 & 0.935 & 0.949 & 0.951 & \second{0.954} \\
   & \textbf{(PMI)}& 0.329 & 0.514 & 0.742 & 0.864 & 0.919 & \best{0.950} & \second{0.953} & 0.953 & \best{0.955} \\
   & \textbf{(weight+unnorm)} & 0.262 & 0.386 & 0.584 & 0.708 & 0.805 & 0.889 & 0.916 & 0.935 & 0.951 \\
   & \textbf{(weight+norm)} & 0.283 & 0.438 & 0.656 & 0.779 & 0.863 & 0.923 & 0.942 & 0.951 & \best{0.955} \\
   & \textbf{(weight+PMI)} & 0.343 & 0.487 & 0.690 & 0.826 & 0.900 & 0.941 & 0.950 & \second{0.954} & \best{0.955} \\
   \hline
  \end{tabular}
  }
}
  \subfigure[AwA2 dataset]{
 \label{table:Accuracy-within-domain-AwA2}
 \centering
 \scalebox{0.8}{
  \begin{tabular}{llcccccccccc}
   \hline
      &  & $k=$1 & 2 & 5 & 10 & 20 & 50 & 100 & 200 & 500 & 1000 \\
   \hline
   \multirow{3}{*}{CDMCA} & ($K=2$) & 0.124 & 0.162 & 0.259 & 0.376 & 0.541 & 0.787 & 0.919 & 0.978 & 0.996 & \best{1.000} \\
   & ($K=200$) & 0.721 & 0.773 & 0.820 & 0.849 & 0.872 & 0.917 & 0.945 & 0.969 & 0.991 & \best{1.000} \\
   & ($K=300$) & 0.730 & 0.787 & 0.837 & 0.859 & 0.887 & 0.916 & 0.947 & 0.970 & 0.992 & \second{0.999} \\
   \cmidrule(lr){1-1}
   \multirow{3}{*}{K-CDMCA} & ($K=2$) & 0.170 & 0.242 & 0.388 & 0.546 & 0.724 & 0.892 & 0.960 & \best{0.991} & 0.998 & \best{1.000} \\
   & ($K=200$) & 0.647 & 0.736 & 0.831 & 0.874 & 0.906 & 0.941 & 0.966 & \second{0.986} & 0.996 & \second{0.999} \\
   & ($K=300$) & 0.661 & 0.748 & 0.833 & 0.877 & 0.904 & 0.942 & 0.967 & \second{0.986} & 0.996 & \best{1.000} \\
   \cmidrule(lr){1-1}
   \multirow{2}{*}{PMvGE-ignored} & ($K=2$) & 0.088 & 0.157 & 0.299 & 0.461 & 0.656 & 0.865 & 0.950 & 0.985 & \second{0.999} & \best{1.000} \\
   & ($K=100$) & 0.038 & 0.058 & 0.059 & 0.060 & 0.060 & 0.080 & 0.100 & 0.140 & 0.260 & 0.460 \\
   \cmidrule(lr){1-1}
   \multirow{2}{*}{PMvGE-2nd-order} & ($K=2$)  & 0.027 & 0.051 & 0.131 & 0.235 & 0.393 & 0.676 & 0.863 & 0.974 & \best{1.000} & \best{1.000} \\
   & ($K=100$) & 0.096 & 0.156 & 0.252 & 0.380 & 0.549 & 0.790 & 0.918 & 0.979 & 0.997 & \best{1.000} \\
   \cmidrule(lr){1-1}
   \multirow{6}{*}{\textbf{MR-SNE}} 
   & \textbf{(unnorm)} & 0.553 & 0.638 & 0.728 & 0.777 & 0.831 & 0.888 & 0.934 & 0.964 & 0.989 & 0.998 \\
   & \textbf{(norm)} & 0.540 & 0.612 & 0.701 & 0.760 & 0.810 & 0.880 & 0.929 & 0.964 & 0.993 & 0.998 \\
   & \textbf{(PMI)} & 0.560 & 0.642 & 0.727 & 0.784 & 0.828 & 0.893 & 0.936 & 0.968 & 0.989 & \second{0.999} \\
   & \textbf{(weight+unnorm)} & \second{0.767} & \second{0.837} & \second{0.899} & \second{0.924} & \second{0.940} & \second{0.958} & \second{0.972} & 0.982 & \second{0.994} & \second{0.999} \\
   & \textbf{(weight+norm)} & \best{0.775} & \best{0.847} & \best{0.904} & \best{0.925} & \best{0.944} & \best{0.963} & \best{0.975} & \second{0.986} & 0.993 & 0.998 \\
   & \textbf{(weight+PMI)} & 0.758 & \second{0.837} & 0.892 & 0.918 & 0.936 & 0.956 & 0.972 & 0.984 & 0.992 & 0.998 \\
   \hline
  \end{tabular}
  }
}
\end{table}
Observation is almost similar to the evaluation by metric (I) across domains. 
For both datasets, overall, MR-SNE demonstrates higher scores than other methods if $K=2$; MR-SNE is expected for providing better visualization. 
If $K>2$, PMvGE demonstrates higher scores than others for Flickr dataset, whereas MR-SNE does for AwA2.

\clearpage
\subsection{Metric (II) across domains}
\label{subsec:metric2_across}

\begin{table}[h]
\centering
\caption{Evaluated by metric (II) across domains. For $K>2$, $t$-SNE is applied to obtain $2$-dim. vectors. Best score is \best{bolded} and the second best is \second{underlined}.}
\label{tab:experiment_d3}
 \subfigure[Flickr dataset]{
\label{table:Number-across-domains-Flickr}
\centering
 \scalebox{0.8}{
  \begin{tabular}{llccccccccc}
   \hline
    &  & $k=$1 & 2 & 5 & 10 & 20 & 50 & 100 & 200 & 500 \\
   \hline
   \multirow{3}{*}{CDMCA} & ($K=2$) & 0.021 & 0.048 & 0.118 & 0.227 & 0.392 & 0.765 & 1.236 & 2.078 & 4.233 \\
   & ($K=200$) & \second{0.758} & \best{1.212} & \best{1.599} & \second{1.694} & 1.798 & 2.081 & 2.538 & 3.322 & 5.090 \\
   & ($K=613$) & \best{0.806} & \second{1.050} & 1.135 & 1.222 & 1.346 & 1.743 & 2.297 & 3.449 & 5.338 \\
   \cmidrule(lr){1-1}
   \multirow{3}{*}{K-CDMCA} & ($K=2$) & 0.024 & 0.033 & 0.071 & 0.151 & 0.359 & 0.780 & 1.214 & 1.916 & 3.967 \\
   & ($K=200$) & 0.046 & 0.074 & 0.147 & 0.217 & 0.321 & 0.648 & 1.099 & 2.075 & 4.503 \\
   & ($K=613$) & 0.059 & 0.087 & 0.160 & 0.260 & 0.423 & 0.776 & 1.312 & 2.171 & 4.616 \\
   \cmidrule(lr){1-1}
   \multirow{2}{*}{PMvGE-ignored} & ($K=2$) & 0.078 & 0.142 & 0.328 & 0.584 & 1.067 & 2.283 & \best{3.672} & \best{4.959} & \best{5.551} \\
   & ($K=100$) & 0.544 & 0.865 & \second{1.357} & \best{1.758} & \best{2.173} & \second{2.823} & 3.467 & 4.323 & \second{5.470} \\
   \cmidrule(lr){1-1}
   \multirow{2}{*}{PMvGE-2nd-order} & ($K=2$) & 0.015 & 0.075 & 0.243 & 0.549 & 1.066 & 1.978 & 2.858 & 3.880 & 5.382 \\
   & ($K=100$) & 0.035 & 0.046 & 0.061 & 0.103 & 0.210 & 0.384 & 0.652 & 1.293 & 4.867 \\
   \cmidrule(lr){1-1}
   \multirow{6}{*}{\textbf{MR-SNE}} 
   & \textbf{(unnorm)} & 0.163 & 0.311 & 0.646 & 1.066 & 1.663 & 2.713 & \second{3.634} & \second{4.590} & 5.417 \\
   & \textbf{(norm)} & 0.234 & 0.426 & 0.801 & 1.286 & 1.873 & \best{2.855} & 3.597 & 4.425 & 5.429 \\
   & \textbf{(PMI)} & 0.319 & 0.549 & 1.069 & 1.519 & \second{2.045} & 2.784 & 3.361 & 4.069 & 5.296 \\
   & \textbf{(weight+unnorm)} & 0.163 & 0.294 & 0.637 & 1.081 & 1.690 & 2.726 & 3.591 & 4.459 & 5.400 \\
   & \textbf{(weight+norm)} & 0.208 & 0.395 & 0.819 & 1.247 & 1.822 & 2.737 & 3.472 & 4.289 & 5.385 \\
   & \textbf{(weight+PMI)} & 0.313 & 0.546 & 0.967 & 1.378 & 1.839 & 2.503 & 3.142 & 3.918 & 5.282 \\
   \hline
  \end{tabular}
  }
}
\subfigure[AwA2 dataset]{
\label{table:Number-across-domains-AwA2}
\centering
 \scalebox{0.8}{
  \begin{tabular}{llcccccccc}
   \hline
       & & $k=$1 & 2 & 3 & 5 & 10 & 20 & 30 & 50 \\
   \hline
   \multirow{3}{*}{CDMCA} & ($K=2$) & 0.579 & 1.142 & 1.728 & 2.892 & 5.766 & 11.169 & 16.257 & 24.929 \\
   & ($K=200$) & 0.246 & 0.591 & 0.855 & 1.491 & 3.340 & 7.269 & 10.922 & 18.555  \\
   & ($K=300$) & 0.264 & 0.636 & 0.973 & 1.582 & 3.110 & 6.849 & 10.906 & 19.327  \\
   \cmidrule(lr){1-1}
   \multirow{3}{*}{K-CDMCA}& ($K=2$) & 0.576 & 1.146 & 1.541 & 2.438 & 4.706 & 8.812 & 12.109 & 19.620  \\
   & ($K=200$) & 0.174 & 0.563 & 1.064 & 1.650 & 3.528 & 7.281 & 11.254 & 19.222  \\
   & ($K=300$) & 0.192 & 0.597 & 1.052 & 1.646 & 3.520 & 7.156 & 11.255 & 19.267  \\
   \cmidrule(lr){1-1}
   \multirow{2}{*}{PMvGE-ignored} & ($K=2$) & \second{0.697} & \second{1.387} & \second{2.072} & \best{3.366} & \best{6.786} & \best{13.547} & \best{19.505} & \best{26.720} \\
   & ($K=100$) & 0.140 & 0.240 & 0.660 & 1.280 & 2.545 & 4.686 & 6.493 & 14.005 \\
   \cmidrule(lr){1-1}
   \multirow{2}{*}{PMvGE-2nd-order} & ($K=2$) & 0.685 & \best{1.442} & \best{2.298} & \second{3.272} & \second{6.694} & \second{12.583} & \second{18.077} & \second{26.386}  \\
   & ($K=100$) & 0.100 & 0.220 & 0.240 & 1.120 & 2.298 & 6.520 & 10.794 & 20.900  \\
   \cmidrule(lr){1-1}
   \multirow{6}{*}{\textbf{MR-SNE}} 
   & \textbf{(unnorm)} & 0.539 & 1.116 & 1.542 & 2.515 & 4.783 & 8.991 & 13.232 & 20.397 \\
   & \textbf{(norm)} & 0.232 & 0.523 & 1.022 & 1.866 & 3.657 & 7.636 & 11.901 & 20.972 \\
   & \textbf{(PMI)} & 0.128 & 0.368 & 0.637 & 1.216 & 2.690 & 6.245 & 11.663 & 22.029 \\
   & \textbf{(weight+unnorm)} & 0.520 & 1.078 & 1.640 & 2.606 & 5.122 & 10.943 & 16.841 & 25.017 \\
   & \textbf{(weight+norm)} & 0.620 & 1.150 & 1.536 & 2.463 & 4.982 & 9.883 & 15.232 & 24.361 \\
   & \textbf{(weight+PMI)} & \best{0.703} & 1.301 & 1.902 & 2.979 & 5.766 & 10.462 & 14.266 & 21.985 \\
   \hline
  \end{tabular}
}
}
\end{table}

Amongst all the methods with $K=2$, for Flickr dataset, MR-SNE demonstrates the best score for $k \le 50$, whereas PMvGE~(ignored) is the best for $k>50$; thus MR-SNE outperforms PMvGE~(ignored) depending on the dataset. 
However, on the other hand, for AwA2 dataset, PMvGE~(ignored) shows better performance than others for $K=2$. 
For $K>2$, PMvGE and CDMCA outperform MR-SNE, whereas increasing $K>2$ does not improve the scores for AwA2 dataset.

\clearpage
\subsection{Metric (II) within image domain}
\label{subsec:metric2_within}

\begin{table}[h]
 \centering
 \caption{Evaluated by metric (II) within image domain. For $K>2$, $t$-SNE is applied to obtain $2$-dim. vectors. Best score is \best{bolded} and the second best is \second{underlined}.}
\label{tab:experiment_d4}
 \subfigure[Flickr dataset]{
\label{table:Number-within-domain-Flickr}
\centering
 \scalebox{0.8}{
  \begin{tabular}{llccccccccc}
   \hline
   &  & $k=$1 & 2 & 5 & 10 & 20 & 50 & 100 & 200 & 500 \\
   \hline
   \multirow{3}{*}{CDMCA} & ($K=2$) & 0.170 & 0.359 & 0.902 & 1.726 & 3.392 & 7.965 & 14.842 & 26.061 & 47.934 \\
   & ($K=200$) & \best{0.887} & \best{1.688} & \best{3.453} & \best{4.801} & 6.102 & 8.941 & 13.375 & 21.465 & 42.486 \\
   & ($K=613$) & \second{0.643} & 1.033 & 1.556 & 2.230 & 3.380 & 6.405 & 11.029 & 19.172 & 40.605 \\
   \cmidrule(lr){1-1}
   \multirow{3}{*}{K-CDMCA} & ($K=2$) & 0.094 & 0.186 & 0.458 & 0.931 & 1.844 & 4.556 & 8.906 & 17.528 & 43.205 \\
   & ($K=200$) & 0.191 & 0.338 & 0.760 & 1.358 & 2.458 & 5.138 & 9.361 & 17.967 & 42.529 \\
   & ($K=613$) & 0.194 & 0.355 & 0.774 & 1.357 & 2.388 & 5.142 & 9.468 & 18.181 & 42.710 \\
   \cmidrule(lr){1-1}
   \multirow{2}{*}{PMvGE-ignored} & ($K=2$) & 0.213 & 0.405 & 0.954 & 1.843 & 3.685 & 8.947 & 16.960 & 31.457 & 62.299 \\
   & ($K=100$) & 0.640 & \second{1.192} &\second{2.563} & \second{4.287} & \best{6.852} & \second{12.690} & \second{19.875} & 31.230 & 56.989 \\
   \cmidrule(lr){1-1}
   \multirow{2}{*}{PMvGE-2nd-order} & ($K=2$) & 0.210 & 0.393 & 0.961 & 1.943 & 3.806 & 9.311 & 18.334 & \second{35.087} & \best{68.956} \\
   & ($K=100$) & 0.641 & 1.147 & 2.389 & 3.963 & \second{6.685} & \best{13.637} & \best{23.434} & \best{37.873} & \second{65.739} \\
   \cmidrule(lr){1-1}
   \multirow{6}{*}{\textbf{MR-SNE}}
   & \textbf{(unnorm)} & 0.266 & 0.549 & 0.954 & 1.843 & 3.685 & 8.947 & 16.960 & 31.457 & 62.299 \\
   & \textbf{(norm)} & 0.296 & 0.581 & 1.414 & 2.721 & 5.045 & 11.036 & 19.525 & 32.592 & 59.104 \\
   & \textbf{(PMI)} & 0.329 & 0.664 & 1.552 & 2.873 & 5.085 & 10.471 & 17.700 & 28.494 & 51.905 \\
   & \textbf{(weight+unnorm)} & 0.262 & 0.497 & 1.233 & 2.424 & 4.640 & 10.791 & 19.479 & 33.216 & 60.680 \\
   & \textbf{(weight+norm)} & 0.283 & 0.558 & 1.350 & 2.544 & 4.775 & 10.461 & 18.302 & 30.544 & 56.870 \\
   & \textbf{(weight+PMI)} & 0.343 & 0.643 & 1.460 & 2.665 & 4.705 & 9.532 & 15.950 & 26.055 & 50.217 \\
   \hline
  \end{tabular}
  }
}
\subfigure[AwA2 dataset]{
\label{table:Number-within-domain-AwA2}
 \scalebox{0.75}{
  \begin{tabular}{llcccccccccc}
   \hline
       &  & $k=$1 & 2 & 5 & 10 & 20 & 50 & 100 & 200 & 500 & 1000 \\
   \hline
   \multirow{3}{*}{CDMCA} & ($K=2$) & 0.124 & 0.178 & 0.312 & 0.548 & 1.020 & 2.4052 & 4.562 & 8.338 & 17.577 & 28.845 \\
   & ($K=200$) & 0.721 & 1.360 & 3.105 & 5.794 & 10.696 & 21.044 & 26.800 & 30.992 & 36.768 & 42.002 \\
   & ($K=300$) & 0.730 & 1.400 & 3.258 & 6.109 & 11.202 & 22.010 & 28.050 & 32.106 & 37.954 & 42.546 \\
   \cmidrule(lr){1-1}
   \multirow{3}{*}{K-CDMCA} & ($K=2$) & 0.170 & 0.270 & 0.554 & 1.003 & 1.788 & 3.702 & 6.210 & 9.869 & 17.610 & 27.447 \\
   & ($K=200$) & 0.647 & 1.188 & 2.486 & 4.070 & 6.018 & 9.304 & 12.810 & 17.692 & 26.788 & 36.288 \\
   & ($K=300$) & 0.661 & 1.218 & 2.546 & 4.181 & 6.130 & 9.384 & 12.755 & 17.562 & 26.443 & 35.956 \\
   \cmidrule(lr){1-1}
   \multirow{2}{*}{PMvGE-ignored} & ($K=2$) & 0.088 & 0.184 & 0.443 & 0.854 & 1.661 & 3.808 & 7.110 & 12.761 & 24.470 & 36.839 \\
   & ($K=100$) & 0.038 & 0.074 & 0.182 & 0.364 & 0.726 & 1.797 & 2.895 & 4.855 & 10.735 & 20.535 \\
   \cmidrule(lr){1-1}
   \multirow{2}{*}{PMvGE-2nd-order} & ($K=2$) & 0.027 & 0.051 & 0.139 & 0.264 & 0.517 & 1.258 & 2.517 & 4.952 & 11.763 & 22.279 \\
   & ($K=100$) & 0.096 & 0.192 & 0.458 & 0.844 & 1.497 & 3.030 & 4.937 & 7.599 & 14.563 & 24.318 \\
   \cmidrule(lr){1-1}
   \multirow{6}{*}{\textbf{MR-SNE}} 
   & \textbf{(unnorm)} & 0.553 & 1.013 & 2.278 & 4.221 & 7.673 & 14.724 & 18.703 & 21.747 & 26.868 & 32.510 \\
   & \textbf{(norm)} & 0.540 & 0.997 & 2.259 & 4.208 & 7.750 & 14.870 & 18.344 & 21.349 & 26.351 & 32.016 \\
   & \textbf{(PMI)} & 0.560 & 1.047 & 2.396 & 4.481 & 8.102 & 15.326 & 19.503 & 22.623 & 26.829 & 32.067 \\
   & \textbf{(weight+unnorm)} & \second{0.767} & \second{1.483} & \second{3.532} & \second{6.824} & \second{13.063} & 27.188 & \second{33.363} & 36.922 & \best{40.957} & \best{44.492} \\
   & \textbf{(weight+norm)} & \best{0.775} & \best{1.492} & \best{3.560} & \best{6.878} & \best{13.227} & \best{27.998} & \best{34.212} & \second{37.498} & 40.739 & \second{43.966} \\
   & \textbf{(weight+PMI)} & 0.758 & 1.472 & 3.502 & 6.777 & 12.948 & \second{27.621} & \best{34.212} & \best{37.804} & \second{40.872} & 43.866 \\
   \hline
  \end{tabular}
}
}
\end{table}

For AwA2 dataset, overall, MR-SNE ourperforms all the other methods, including those for $K>2$; thus MR-SNE is expected to provide good visualization. 
For Flickr dataset, MR-SNE shows the best score for $K=2$, whereas PMvGE outperforms MR-SNE for $K>2$. 
This is also the same observation as the experiments in Supplement~\ref{subsec:metric1_across}.

\clearpage
\section{PMvGE using Stochastic Neighbor Graph}
\label{supp:stochastic_pmvge}

Similarly to the proposed MR-SNE, we may consider employing stochastic neighbor~(SN) graph for PMvGE. 
Although such an augmented PMvGE using SN graph is not included in the original PMvGE, meaning that this extension may not be included in baselines, we here perform this augmented PMvGE to be compared with MR-SNE. 

\begin{figure}[H]
\centering
\subfigure[Flickr]{
		\includegraphics[width=5cm]{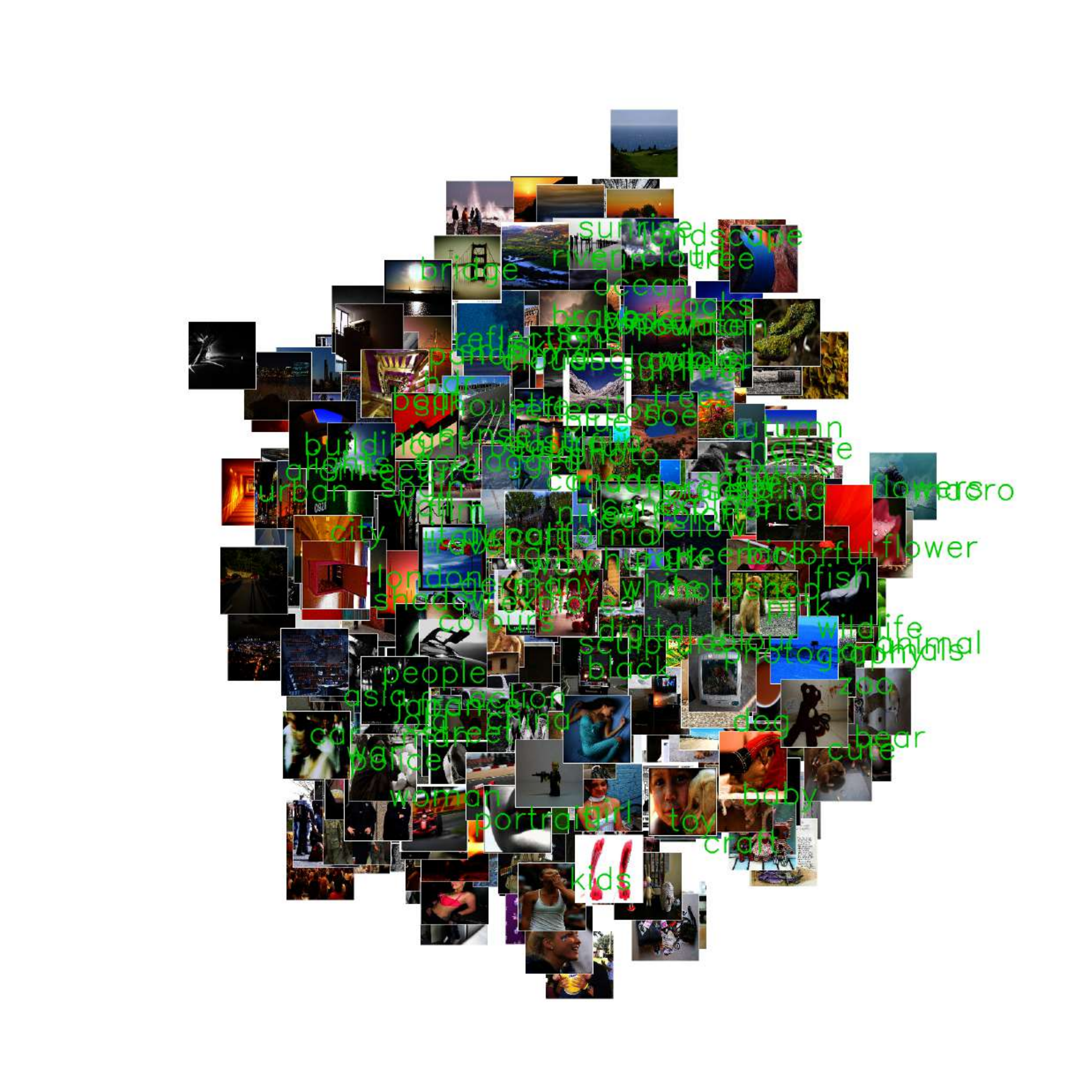}
	\label{fig:Flickr_Stochastic}
}
\hspace{2em}
\subfigure[AwA2]{
	\includegraphics[width=5cm]{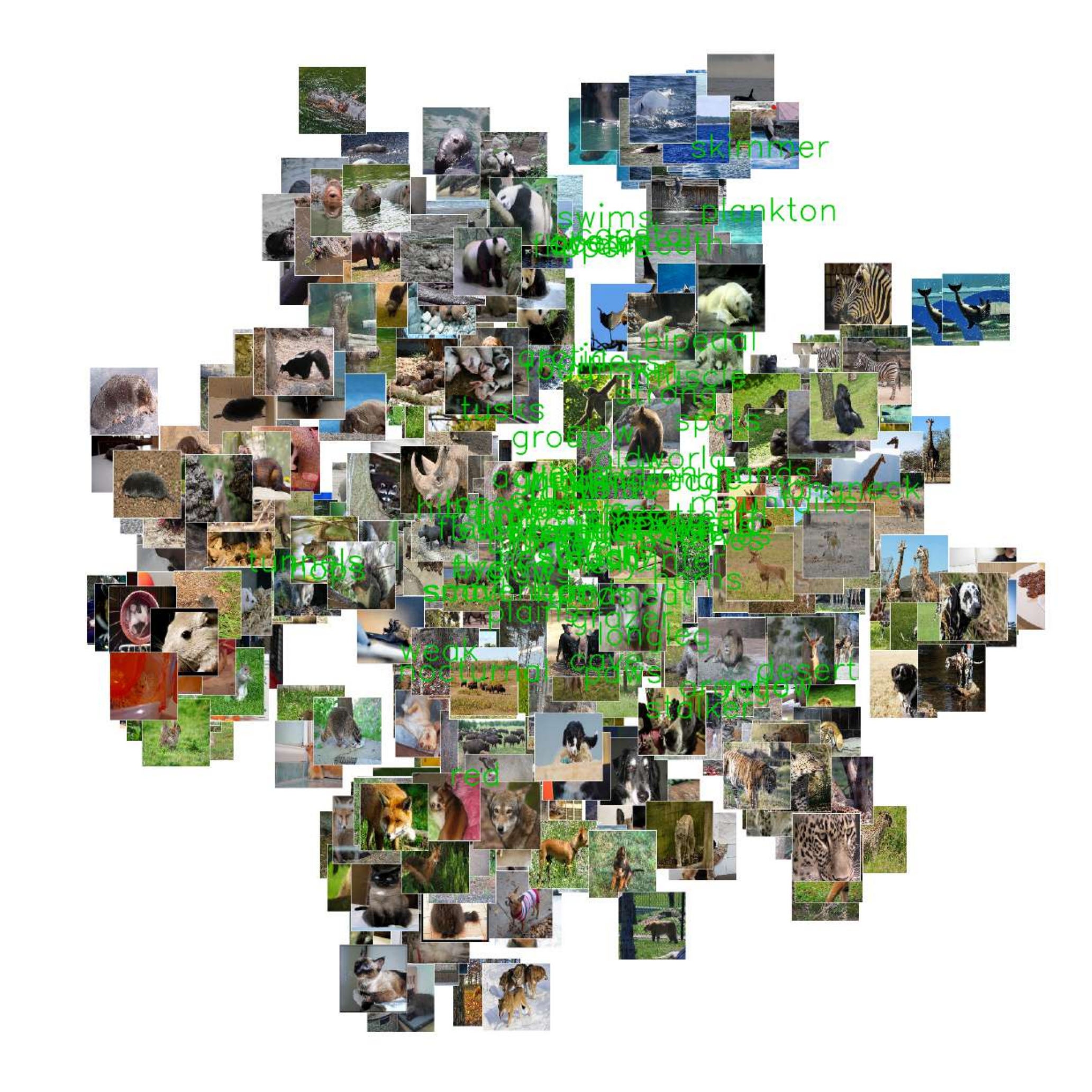}
	\label{fig:AwA2_Stochastic}
}
\caption{Visualization of PMvGE using stochastic neighbor graph.}
\label{fig:Visualization_Stochastic}
\end{figure}

For training the PMvGE, we sampled $10^5$ edges with probabilities depending on the SN graph, that we call as ``positive" edges; using a graph consisting of the sampled edges, we optimize PMvGE using negative sampling. Then, Flickr and AwA2 datasets are visualized in Figure~\ref{fig:Visualization_Stochastic}. 

ROC-AUC scores and variance scores considered in Section~\ref{subsec:experiment2}, \ref{subsec:experiment3} are shown in the following Table~\ref{tab:Stochastic}.

\begin{table}[htbp]
\centering
\caption{ROC-AUC and variance scores are listed; $t$-SNE is applied to obtain $2$-dimensional vectors for $K>2$. Best score is \best{bolded} and the second best is \second{underlined}.}
\label{tab:Stochastic}
    \begin{tabular}{llcccc}
    \toprule[0.2ex]
         & &  \multicolumn{2}{c}{ROC-AUC} & \multicolumn{2}{c}{Variance score} \\
         & &  Flickr & AwA2 & Flickr & AwA2\\
        \cmidrule(lr){1-2} \cmidrule(lr){3-4} \cmidrule(lr){5-6}
        \multirow{6}{*}{MR-SNE} 
        & (unnorm) & \best{0.7134} & 0.7117 & 1.706 & 56.563\\
        & (norm) & \second{0.7038} & 0.7161 & 1.388 & 26.139 \\
        & (PMI) & 0.6690 & 0.7109 & 1.151 & 27.800 \\
         & (weight+unnorm) & 0.7000 & 0.8033 & 1.556 & \best{1.087}\\
        & (weight+norm) & 0.6902 & 0.8124 & 1.475 & 1.190 \\
        & (weight+PMI) & 0.6542 & 0.8134 & \second{1.067} & \second{1.165} \\
        \cmidrule(lr){1-1}
        \multirow{2}{*}{PMvGE + SN graph} & $(K=2)$ & 0.6966 & \second{0.8588} & 1.143 & 0.042  \\
         & $(K=100)$ & 0.6717 & \best{0.8863} & \best{1.064} & 2.670 \\
        \bottomrule[0.2ex]
    \end{tabular}
\end{table}
Overall, PMvGE using SN graph demonstrates good ROC-AUC scores; the score is better than MR-SNE for AwA2, though worse for Flickr dataset. 
Similarly to most of other baselines, variance score is far from $1$ for AwA2 dataset; MR-SNE with proper weights is yet more stable than the PMvGE using SN graph.

\clearpage
\section{CDMCA and Kernel CDMCA}
\label{supp:cdmca}

Canonical Correlation Analysis~\citep[CCA;][]{CCA} is one of the most well-known and widely-accepted methods for multi-view data analysis. However, CCA is designed for considering one-to-one correspondence across domains, meaning that across-domains graph cannot be considered in CCA; Cross-Domain Matching Correlation Analysis~\citep[CDMCA;][]{CDMCA} extends CCA so that the across-domains graph is considered. 
In fact, a special case of (kernel) CDMCA for $2$ domains is obtained by applying (kernel) CCA to preprocessed data vectors; the procedure is shown in the following Section~\ref{subsec:preprocessing}. 

For implementation of CCA and kernel CCA, we employ Pyrcca~\citep{pyrcca}, that is an open-source Python package for kernel CCA; therein, we specify the regularization parameter by $\lambda =0.01$.

\subsection{Preprocessing}
\label{subsec:preprocessing}
\begin{figure}[htbp]
\centering
\includegraphics[width=0.80\textwidth]{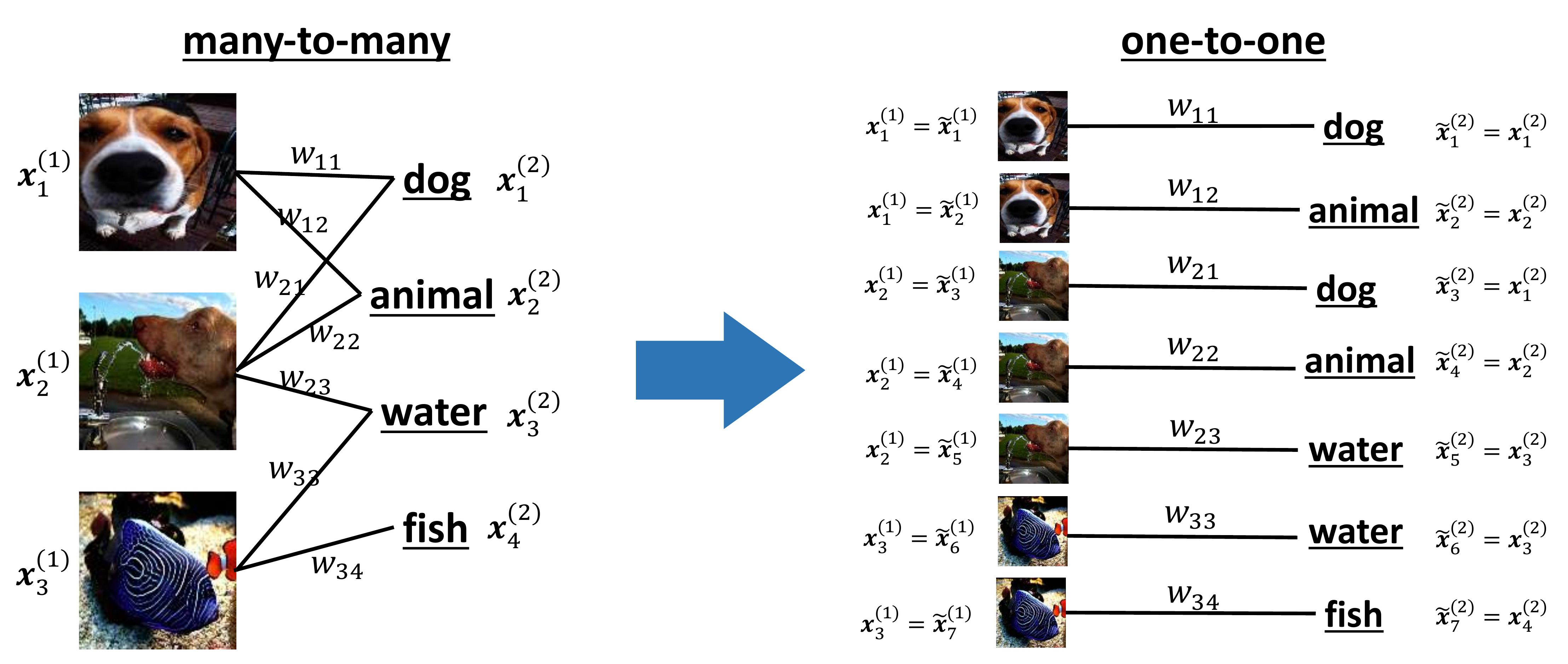}
\caption{Many to many relations can be converted to one-to-one relations by duplicating some data vectors; applying CCA to the new data vectors yields CDMCA ($D=2$).}
\label{fig:cdmca_image}
\end{figure}

Assuming that an across-domains graph $W=(w_{ij}) \in \{0,1\}^{n_1 \times n_2}$ is observed. 
Note that, we here consider only the binary weights $w_{ij} \in \{0,1\}$ representing links in the graph, and within-domain graphs are not considered. 
Then, we first compute a set of index pairs whose corresponding weights are positive, i.e., $\mathcal{S}:=\{(i,j) \mid 1 \le i \le n_1,1 \le j \le n_2, w_{ij}=1\}$; by assigning new symbols, the set can also be written as
\[
    \mathcal{S}=\{(i_1,j_1),(i_2,j_2),\ldots,(i_m,j_m)\},
\]
where $m \in \mathbb{N}$ is the number of links in the graph.
Then, by specifying new data vectors 
\begin{align}
\tilde{x}^{(1)}_{k}=x^{(1)}_{i_k}, \quad
\tilde{x}^{(2)}_{k}=x^{(2)}_{j_k}, \quad 
(k=1,2,\ldots,m),
\label{eq:preprocessed_data_vectors}
\end{align}
the two data vectors in the pair $(\tilde{x}^{(1)}_{k},\tilde{x}^{(2)}_k)$
are associated by the link;
we have copied the same data vectors several times when they have multiple links.
Now, the preproccessed new data vectors $\{\tilde{x}^{(1)}_k\}_{k=1}^{m},\{\tilde{x}^{(2)}_k\}_{k=1}^{m}$ have one-to-one association with positive weight $w_{i_kj_k}=1$ as illustrated in Figure~\ref{fig:cdmca_image}. 
Applying CCA to the preprocessed new data vectors (\ref{eq:preprocessed_data_vectors}) is equivalent to a special case of CDMCA for $2$-domains; kernel CDMCA is also similarly defined, by applying kernel CDMCA~\citep{KCCA} to the data vectors~(\ref{eq:preprocessed_data_vectors}).

\end{document}